\documentclass[11pt]{article}

\usepackage[preprint]{acl}

\usepackage{times}
\usepackage{latexsym}
\usepackage{booktabs}
\usepackage{amsmath}
\usepackage{amssymb}

\usepackage[T1]{fontenc}

\usepackage[utf8]{inputenc}

\usepackage{microtype}

\usepackage{inconsolata}

\usepackage{graphicx}
\usepackage{xcolor}
\usepackage{tikz}
\usetikzlibrary{positioning, arrows.meta, calc}
\usepackage{adjustbox}
\usepackage{caption}
\usepackage{cuted}
\usepackage{pgfplots}
\usepackage{xurl}

\definecolor{teaserInput}{RGB}{230,238,247}   %
\definecolor{teaserSingle}{RGB}{225,228,232}  %
\definecolor{teaserJa}{RGB}{205,232,224}      %
\definecolor{teaserJb}{RGB}{248,225,202}      %
\definecolor{teaserJc}{RGB}{226,215,238}      %
\definecolor{teaserAgg}{RGB}{255,220,150}     %
\definecolor{teaserReward}{RGB}{120,40,90}    %

\title{Judging the Judges: Confidence and Disagreement in Rubrics as Rewards}

\author{Michael Tian-Yue Liu\thanks{Equal contribution.} \and Anne Harrington\footnotemark[1] \and Trevor Darrell \and Yutong Bai \\
UC Berkeley \\
\texttt{\{liu.michael, anneko, trevordarrell, yutongbai\}@berkeley.edu}}

\begin{document}
\maketitle

\begin{abstract}

Rubrics as Rewards (RaR) extends reinforcement learning to non-verifiable domains by using an LLM judge to score responses against a structured rubric. Prior work has focused on improving the rubric or strengthening the single judge, with comparatively little attention to how different single or multiple judges might be used during RaR training. We first show that judge identity alone produces a substantial spread in scores when training Qwen3-4B-Base on RaR-Medicine and RaR-Science across five frontier and open-weight judges. Building on this, we propose \textbf{RaRE} (Rubrics as Rewards with Ensembles), a novel judge ensemble strategy for RaR with two mechanisms: a \emph{soft-Likert reward} that uses judge logprobs to weight more confident judges more heavily, and a \emph{drop-on-disagreement mask} that excludes high-disagreement rollouts from the GRPO update. We find that this design is critical: while a naive average of judge scores fails to outperform the best single judge, RaRE outperforms every single-judge RaR variant on both HealthBench and GPQA Diamond, with per-theme analysis showing that it inherits the complementary strengths of its constituent judges. Our findings establish judge ensembling as a compelling axis for improving RaR.\footnote{Project page:  \url{https://anneharrington.github.io/judging-the-judges/}}
\end{abstract}

\section{Introduction}

Rubrics as Rewards (RaR; \citealp{gunjal2025rubrics}) is increasingly used in LLM post-training for learning on non-verifiable domains. Unlike Reinforcement Learning with Verifiable Rewards (RLVR), which is standard for math and code \citep{lambert2024tulu,guo2025deepseekr1}, RaR enables training in more realistic settings where response quality might depend on multiple criteria that are not easily checked (such as completeness, style, and safety). RaR consists of two components: a structured, instance-specific rubric that encodes relevant criteria, and an LLM judge that scores each rollout against the rubric to produce a scalar reward. Following its success, subsequent work has explored in depth how to improve rubrics themselves; these include better offline construction \citep{zhang2025chasing,liu2025openrubrics}, checklist-style scaffolding during rollouts \citep{zhou2025breaking}, and online rubric updates \citep{rezaei2025online,li2026evolm}.  While the effect of the judge has been studied extensively at evaluation time, including work on judge ensembles \citep{verga2024replacing,chan2024chateval,feng2025we}, comparatively the impact of the judge during RaR training has been under-explored.

To fill this gap, we evaluate the role of the judge model during RaR training more broadly. We train Qwen3-4B-Base \citep{yang2025qwen3} on RaR-Medicine and RaR-Science with the same strategy and rubrics as \citet{gunjal2025rubrics} across 5 open and closed weight judging models. We show that varying judge identity alone in RaR produces a substantial spread in scores across both domains.

Using our insights on the effect of the single judge model, we then develop \textbf{RaRE} (Rubrics as Rewards with Ensembles), a novel ensemble strategy to improve RaR. Rather than taking a naive average of judge scores, which we show fails to outperform the best single judge, RaRE targets two distinct sources of uncertainty. First, we introduce a \emph{soft-Likert reward} to leverage intra-judge confidence: we use each judge's logprobs to weight more confident judges more heavily in the ensemble. Second, we incorporate a \emph{drop-on-disagreement mask} to leverage inter-judge disagreement: we exclude rollouts on which judges strongly disagree from the GRPO update. The two mechanisms can be used independently or composed.

We validate RaRE on an ensemble of two judges, testing on the medicine and science domains using HealthBench \citep{arora2025healthbench} and GPQA Diamond \citep{rein2023gpqagraduatelevelgoogleproofqa} respectively. We find that our best RaRE configuration outperforms every single-judge RaR variant in both domains (0.364 vs.\ 0.339 best single judge on HealthBench; 0.412 vs.\ 0.406 best single judge on GPQA Diamond). Examining results by HealthBench theme, we find that RaRE consistently inherits the strengths of its judges, winning on all seven themes.

\begin{figure*}[t]
\centering
\includegraphics[width=\textwidth]{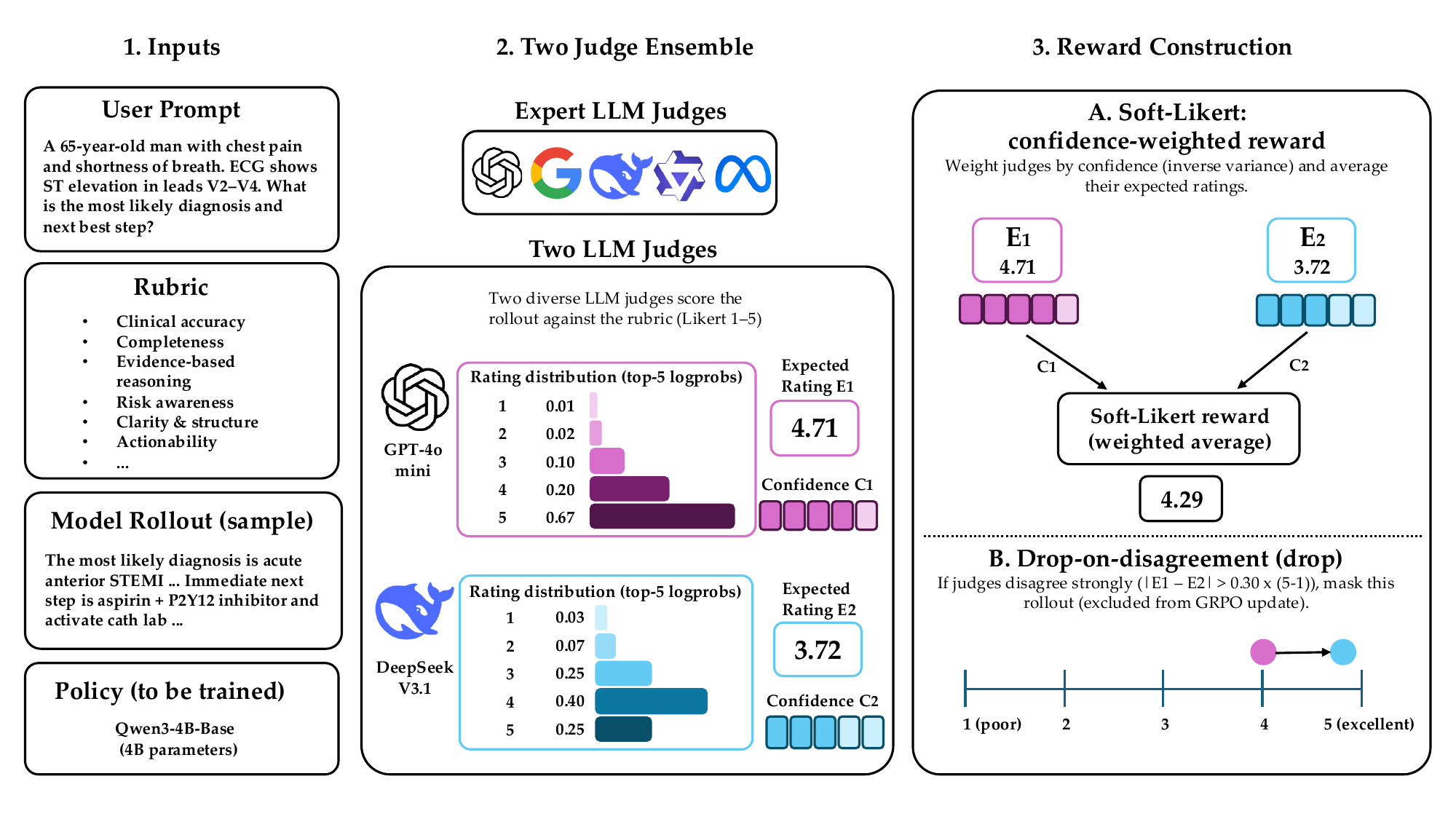}
\caption{Overview of RaRE (Rubrics as Rewards with Ensembles), instantiated with two judges. \textbf{(1) Inputs:} the user prompt, rubric, and a model rollout sampled from the policy (Qwen3-4B-Base) are passed to the judges. \textbf{(2) Two-judge ensemble:} we use an ensemble of two diverse LLM judges (selected from a pool of frontier and open-weight models). Each judge scores the rollout against the rubric on a 1--5 Likert scale, and we read its top-5 logprobs to recover an expected rating $E_i$ and a per-judge confidence $C_i$. \textbf{(3) Reward construction:} our \emph{soft-Likert reward} (A) combines the two expected ratings into a single scalar by weighting each judge by its inverse-variance confidence. Our \emph{drop-on-disagreement mask} (B) excludes rollouts where the two judges' expected ratings differ by more than $\tau \cdot (5-1)$ from the GRPO update.}
\label{fig:teaser}
\end{figure*}

Together, our results suggest that the judge model is a major design choice in RaR, and that small ensembles with the right aggregation can deliver a stronger training signal than any single judge. Our main contributions are as follows:
\begin{enumerate}
    \item A study of the judge in RaR training across five frontier and open-weight judges on RaR-Medicine and RaR-Science, showing that judge identity alone substantially shifts policy quality.
    \item RaRE, a novel ensemble method for RaR with two composable mechanisms: a \emph{soft-Likert reward} that reads judge logprobs to recover confidence-weighted ratings, and a \emph{drop-on-disagreement mask} that excludes high-disagreement rollouts from the GRPO update.
    \item Empirical results showing that RaRE outperforms single-judge RaR variants in both medicine and science, with per-theme analysis revealing that it inherits complementary strengths of its judges, unlike a naive ensemble approach.
\end{enumerate}

\section{Related Work}

\paragraph{RL with verifiable rewards.} Recent gains on math and code
use deterministic checks (e.g., calculators, compilers) as the reward
signal \citep{lambert2024tulu, guo2025deepseekr1}, typically optimized
with GRPO \citep{shao2024deepseekmath}. These checks do not exist for
open-ended domains such as medical or scientific~\citep{su2025crossing}.

\paragraph{Rubric-based rewards.} Rubrics-as-Rewards
\citep{gunjal2025rubrics} and similar work
\citep{viswanathan2026checklists} replace the binary verifier with an
LLM judge that scores a candidate against a structured rubric,
converting a fuzzy preference target into a scalar reward. Subsequent
work has focused almost entirely on the \emph{rubric} side: better
offline construction \citep{zhang2025chasing, liu2025openrubrics},
checklist-style scaffolding during rollouts
\citep{zhou2025breaking}, online rubric updates
\citep{rezaei2025online, li2026evolm}, and causally grounded rubrics
for robust reward modeling \citep{ICLR2026_fa0a1013}. Concurrent work \citep{mahmoud2026rewardhackingrubricbasedreinforcement} also uses multiple judges in the RaR setting, but as a multi-judge evaluation panel for detecting reward hacking, while RaRE uses the judge ensemble as the training signal itself. Across these works, the training judge model is
consistently treated as a fixed implementation detail, usually a
single model. Our work holds the rubric fixed and treats
the judge itself as the design variable, complementary to all of the
above.

\paragraph{LLM-as-a-judge and verification.} LLM-as-a-judge
\citep{zheng2023judging} has become the dominant evaluation paradigm,
with growing emphasis on structured or rubric-conditioned grading
\citep{kim2024prometheus, hashemi2024llm}. Recent reliability studies
\citep{lu2025does, singh2026v_1, kwok2026llmverifier, feng2025we}
show that single judges exhibit substantial position bias, length
bias, and inter-seed variance. Our results in
\S\ref{sec:judge-matters} extend this diagnosis from evaluation to
optimization: judge choice meaningfully affects the trained policy
even with a fixed rubric.

\paragraph{Judge ensembles.} The closest prior work is
Panel-of-LLM-evaluators \citep{verga2024replacing}, which mean-pools
or majority-votes across a panel of smaller judges and shows that the
panel outperforms a single large judge on evaluation while reducing
intra-model bias. ChatEval \citep{chan2024chateval} explores a
debate-based variant, and similar work \citep{feng2025we} finds
that panel-based judging improves preference consistency. To our
knowledge, all of this literature treats judge ensembles as an
\emph{evaluation-time} mechanism. 
RaRE instead uses a judge ensemble inside the RL training loop as the reward signal, with mechanisms (\S\ref{sec:soft-likert-def}, \S\ref{sec:drop-disagreement-def}) that use inter-judge spread as part of the training signal rather than only as a diagnostic.

\section{Methods}
\label{sec:methods}

\paragraph{Notation.} Let $x$ be a prompt, $\pi_\theta$ a policy with parameters $\theta$, and
$y \sim \pi_\theta(\cdot \mid x)$ a sampled response. Following
\citet{gunjal2025rubrics}, a rubric for $x$ is a set of $k(x)$ weighted
criteria
$
R(x) = \{(w_j, c_j)\}_{j=1}^{k(x)},
$
where each $c_j$ is a natural-language criterion paired with a binary
correctness check $c_j(x, y) \in \{0,1\}$ and weight
$w_j \in \mathbb{R}$.
We use implicit aggregation, where the judge consumes the whole rubric and
emits a single scalar, $J(y, x, R) \;\in\; [0,1], $ typically via a Likert score that is min--max normalized
\citep{gunjal2025rubrics}, which subsumes
RLVR \citep{gunjal2025rubrics} as the $k{=}1$, binary case.

\paragraph{Single-judge RaR.}
Single-judge RaR optimizes the KL-regularized reward objective
\citep{zhang2025chasing},
\begin{align}
\label{eq:rar-objective}
\mathcal{J}(\theta)
&= \mathbb{E}_{x \sim \mathcal{D},\, y \sim \pi_\theta(\cdot \mid x)}
   \bigl[ J(y, x, R(x)) \bigr] \notag \\
&\quad - \beta\, D_{\mathrm{KL}}\!\bigl[\pi_\theta \,\big\|\, \pi_{\mathrm{ref}}\bigr].
\end{align}
with policy updates from GRPO, following \citet{gunjal2025rubrics}
and \citet{zhang2025chasing}.

\subsection{Rubrics as Rewards with Ensembles}

We propose \textbf{RaRE} (Rubrics as Rewards with Ensembles), which
replaces the single training judge with an ensemble of $K$ judges. We
first describe the naive ensemble baseline
(\S\ref{sec:naive-ensemble-def}), then RaRE's two composable
mechanisms: the soft-Likert reward (\S\ref{sec:soft-likert-def}) and
the drop-on-disagreement mask (\S\ref{sec:drop-disagreement-def}).

\subsubsection{Naive judge ensemble}
\label{sec:naive-ensemble-def}

Given $K$ judges $\{J_k\}_{k=1}^K$, the naive ensemble reward is the
mean of the individual judge scores:
\begin{equation}
\label{eq:naive-ensemble}
r_{\mathrm{naive}}(y, x, R)
= \frac{1}{K} \sum_{k=1}^{K} J_k(y, x, R).
\end{equation}
This averages out idiosyncratic per-judge bias while keeping the
training pipeline unchanged.

\subsubsection{Soft-Likert reward via judge logprobs}
\label{sec:soft-likert-def}

The naive ensemble (\S\ref{sec:naive-ensemble-def}) treats each
judge's rating as a single integer, discarding the judge's
\emph{confidence} in that rating. We propose a soft-Likert reward
that recovers confidence from the judge's output distribution
and uses it both as a continuous reward and as a weight for ensemble
aggregation.

\paragraph{Setup.} Following \citet{gunjal2025rubrics}, each judge $j$
emits an integer rating on a $1, \dots, L$ Likert scale via strict
JSON-schema decoding: \texttt{\{"rating": $\hat{k}$\}}, where the
rating digit is a single token in the output. Standard implicit-RaR
reads only the emitted integer $\hat{k}$. Instead, we read the
top-$T$ logprobs at the rating-digit position to recover the full
distribution over ratings. Because each rating must fit in one token,
we use $L=T=5$ for soft-Likert judges; the standard $L=10$ scale is
retained for the hard-Likert fallback described below.

\paragraph{Per-judge distribution.} Let $\ell_j^{(i)}(k)$ be the
log-probability the judge assigns to digit $k$ at the rating-digit
position for rollout $i$. The top-$T=5$ logprobs may include non-digit
tokens, so $\sum_{k=1}^L \exp(\ell_j^{(i)}(k)) < 1$ in general. We
exponentiate and renormalize over digits:
\begin{equation}
\label{eq:p-renorm}
p_j^{(i)}(k) =
\frac{\exp(\ell_j^{(i)}(k))}
     {\sum_{k'=1}^{L} \exp(\ell_j^{(i)}(k'))},
\quad k \in \{1, \dots, L\}.
\end{equation}

\paragraph{Expected rating and confidence.} From $p_j^{(i)}$ we
compute the expected rating and an inverse-variance confidence:
\begin{align}
E_j^{(i)} &= \textstyle\sum_{k=1}^{L} k \cdot p_j^{(i)}(k),
   \label{eq:e-rating} \\
\mathrm{Var}_j^{(i)} &= \textstyle\sum_{k=1}^{L} (k - E_j^{(i)})^2 p_j^{(i)}(k),
   \label{eq:var-rating} \\
c_j^{(i)} &= \frac{1}{1 + \mathrm{Var}_j^{(i)}}.
   \label{eq:conf}
\end{align}
$E_j^{(i)} \in [1, L]$ is a continuous version of the integer rating;
$c_j^{(i)} \in (0, 1]$ is $1$ when the judge places all mass on a
single digit, and decreases as the distribution spreads or
bimodalizes. The $+1$ smoothing keeps $c_j \le 1$ and avoids
division by zero. 
We compare alternative confidence schemes
(top-logit, winner-takes-all) in
\S\ref{sec:disagreement-ensemble} and find inverse-variance is the
strongest.

\paragraph{Confidence-weighted ensemble.} The soft-Likert ensemble
reward for rollout $i$ is the confidence-weighted average expected
rating across $K$ judges, scaled to $[1/L, 1]$ by dividing by $L$:
\begin{equation}
\label{eq:soft-likert-ensemble}
r_{\mathrm{soft}}^{(i)}
 = \frac{1}{L} \cdot
 \frac{\sum_{j=1}^{K} c_j^{(i)} E_j^{(i)}}
      {\sum_{j=1}^{K} c_j^{(i)}}.
\end{equation}
This is inspired by the maximum-likelihood combination of independent estimators
under Gaussian likelihoods; for the bounded Likert distribution it is
a soft approximation.

\paragraph{Hard-Likert fallback.} If a judge endpoint does not return logprobs, we parse the integer rating $\hat{k}$ from
the output and set $p_j(k) = \mathbf{1}[k = \hat{k}]$, giving
$\mathrm{Var}_j = 0$ and $c_j = 1$. The ensemble reduces to a plain
average, which we call \emph{hard-Likert}.

\subsubsection{Drop on disagreement}
\label{sec:drop-disagreement-def}

Rollouts where the two judges strongly disagree produce unreliable 
reward estimates, where averaging does not reflect either judge's actual assessment.
We propose \emph{drop-on-disagreement}, a reward-substitution mechanism that
removes these rollouts from the GRPO update by replacing their reward
with the group mean, so that their group-normalized advantage is zero.

\paragraph{Disagreement statistic.} For rollout $i$, let
$\hat{r}_j^{(i)}$ denote judge $j$'s rating.
We measure inter-judge
disagreement by the normalized rating spread,
\begin{equation}
\label{eq:disagreement}
d^{(i)} = \frac{\max_{j} \hat{r}_j^{(i)} - \min_{j} \hat{r}_j^{(i)}}{L - 1}
\;\in\; [0, 1].
\end{equation}

\paragraph{Group-mean substitution.} 
For group $g$ with rollout
indices $\mathcal{I}_g$ and threshold $\tau$, let $\mu^{(i)} =
\mathbf{1}[d^{(i)} > \tau]$ mask the high-disagreement rollouts, and
let $r^{(i)}$ denote the per-rollout reward ($r_{\mathrm{soft}}^{(i)}$
under soft-Likert, or the naive ensemble mean under hard-Likert). We
replace the masked rollouts' rewards with the unmasked group mean,
\begin{align}
\bar{r}_g
 &= \frac{\sum_{i \in \mathcal{I}_g} (1-\mu^{(i)})\, r^{(i)}}
              {\sum_{i \in \mathcal{I}_g} (1-\mu^{(i)})},
 \label{eq:group-mean} \\
r_{\mathrm{drop}}^{(i)}
 &= \begin{cases}
   \bar{r}_g & \text{if } \mu^{(i)} = 1, \\
   r^{(i)} & \text{otherwise.}
 \end{cases}
 \label{eq:drop-reward}
\end{align}
Under GRPO's group-normalized advantage computed on the substituted rewards, $A^{(i)} = (r^{(i)}_{\mathrm{drop}} -
\bar{r}_g) / \sigma_g$, so masked rollouts receive $A^{(i)} = 0$
and contribute no gradient.

\paragraph{Composing confidence and disagreement.} Soft-Likert's inverse-variance
confidence weighting handles \emph{intra}-judge uncertainty, while drop-on-disagreement handles
\emph{inter}-judge uncertainty. They
compose orthogonally (Eqs.~\eqref{eq:soft-likert-ensemble} and
\eqref{eq:drop-reward} can be applied together), and we show in
\S\ref{sec:disagreement-ensemble} that the composition outperforms
either mechanism alone.

\begin{table}[t]
\centering
\footnotesize
\setlength{\tabcolsep}{3pt}
\begin{tabular}{@{}lcccc@{}}
\toprule
Method & Rubric & Ensemble & Confid. & Disag. \\
\midrule
RLVR                & $\times$   & $\times$   & $\times$   & $\times$        \\
RaR  & \checkmark & $\times$   & $\times$   & $\times$    \\
\midrule
Naive ensemble      & \checkmark & \checkmark & $\times$   & $\times$    \\
\quad + drop        & \checkmark & \checkmark & $\times$   & \checkmark  \\
\quad + soft-Likert & \checkmark & \checkmark & \checkmark & $\times$    \\
\midrule
\textbf{RaRE} (ours) & \checkmark & \checkmark & \checkmark & \checkmark  \\
\bottomrule
\end{tabular}
\caption{Comparison of methods.
\textbf{Rubric:} reward comes from an LLM judge scoring a structured
rubric.
\textbf{Ensemble:} reward aggregates ensemble of $K{>}1$ judges
(\S\ref{sec:naive-ensemble-def}).
\textbf{Confid.:} soft-Likert inverse-variance confidence
weighting (\S\ref{sec:soft-likert-def}).
\textbf{Disag.:} drop-on-disagreement mask
(\S\ref{sec:drop-disagreement-def}).}
\label{tab:method-matrix}
\end{table}

\section{Results}

\subsection{Experiment Setup}
\label{sec:setup}

\paragraph{Policy and reference models.} Our trainable policy is
Qwen3-4B-Base \citep{yang2025qwen3}; we report Qwen3-4B as a tuned reference throughout.

\paragraph{Judges.} Five candidate single-judge configurations span
proprietary, open-weight, and small-model regimes: gpt-4o-mini \citep{openai2024gpt4omini},
DeepSeek-V3.1 \citep{deepseekai2024deepseekv3}, Gemini-2.5-flash-lite \citep{comanici2025gemini25} , 
Qwen3-235B-Instruct \citep{yang2025qwen3}, and gpt-oss-120b \citep{openai2025gptoss}.
All judges share the same rubric and same Likert-rating prompt template
(Appendix~\ref{app:judge-prompts}). Soft-Likert
(Eq.~\ref{eq:soft-likert-ensemble}) requires the endpoint to return
\texttt{top\_logprobs}. We queried logprobs for the two
judges used in our soft-Likert ensemble experiments (gpt-4o-mini and
DeepSeek-V3.1); Gemini does not expose this in its API. All judges were run with no thinking or low effort reasoning when applicable.

\paragraph{Training.} We train on RaR-Medicine and RaR-Science \citep{gunjal2025rubrics}
with GRPO \citep{shao2024deepseekmath} using $G{=}8$ generations per
prompt and per-group reward normalization. Full hyperparameters stated
in Appendix~\ref{app:hyperparams}. Headline numbers use full-parameter
fine-tuning; we report a brief LoRA comparison in the length-hacking
discussion below.

\paragraph{Evaluation.} We evaluate on the English-only subset of 
HealthBench for medicine, using gpt-4o-mini as the eval judge \citep{arora2025healthbench}, and GPQA Diamond for science \citep{rein2023gpqagraduatelevelgoogleproofqa}, with max completion tokens capped at 8192 (responses are truncated past this). Throughout, we refer to the English-only subset simply as HealthBench. Furthermore, we split GPQA Diamond into disjoint 50/50 validation/test splits for purposes of hyperparameter tuning, with results reported on the test split.
For some ablations, we evaluate on a subset of 1{,}000 prompts 
(randomly sampled, stratified by theme) which we refer to as the HealthBench 1K subset (with max completion tokens capped at 1024).
We report the
mean HealthBench/GPQA Diamond score (henceforth ``score'') across prompts. For GPQA Diamond, we train models on 5 seeds and report the mean score and 95\% confidence intervals.
HealthBench provides bootstrapped 95\% confidence intervals, stated in Appendix ~\ref{app:confidence-intervals}. 

\subsection{Single-Judge RaR is Sensitive to Judge}
\label{sec:judge-matters}

\begin{figure}[t]
\centering
\includegraphics[width=\columnwidth]{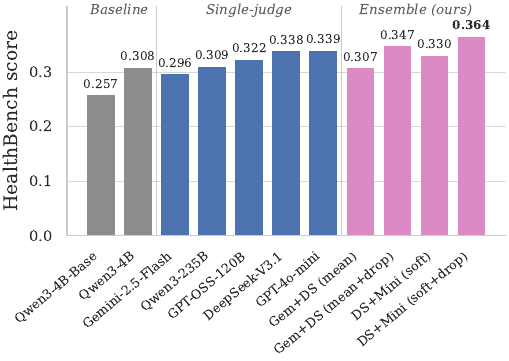}
\caption{HealthBench scores under RaR training across baselines (Qwen3-4B-Base, Qwen3-4B), five single-judge variants, and our RaRE ensembles. \texttt{Gem+DS} pairs Gemini-2.5-Flash-lite with DeepSeek-V3.1; \texttt{DS+Mini} pairs DeepSeek-V3.1 with GPT-4o-mini. Ensemble configurations: \emph{mean} averages judge ratings; \emph{soft} applies our soft-Likert reward; \emph{drop} applies our drop-on-disagreement mask ($\tau=0.30$). Best result in bold.}
\label{fig:healthbench-bar}
\end{figure}

\begin{figure}[t]
\centering
\includegraphics[width=\columnwidth]{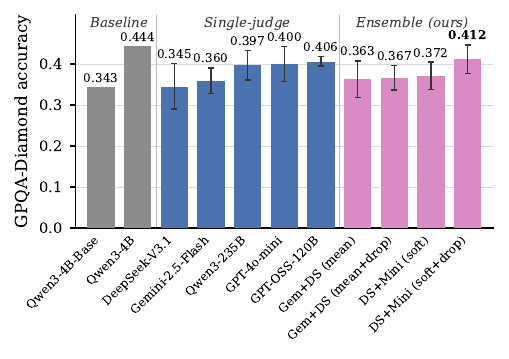}
\caption{GPQA Diamond accuracy on a 50\% test split under RaR training across baselines (Qwen3-4B-Base, Qwen3-4B), five single-judge variants, and our RaRE ensembles. Error bars show the mean over five training seeds with 95\% confidence intervals. \texttt{Gem+DS} pairs Gemini-2.5-Flash-lite with DeepSeek-V3.1; \texttt{DS+Mini} pairs DeepSeek-V3.1 with GPT-4o-mini. Ensemble configurations: \emph{mean}
averages judge ratings; \emph{soft} applies our soft-Likert reward;
\emph{drop} applies our drop-on-disagreement mask ($\tau=0.30$ for
\texttt{Gem+DS}; $\tau=0.40$ for \texttt{DS+Mini}, selected on a held-out
validation set). Best RaR-trained result in bold.}
\label{fig:gpqa-bar}
\end{figure}

Figure~\ref{fig:healthbench-bar} compares RaR-trained Qwen3-4B-Base under five
different judges, all sharing the same rubric (training dynamics in
Fig.~\ref{fig:reward-train}). The spread across
judges ($0.296$--$0.339$ on English-only HealthBench, $\sim$4.3\,pp) is comparable
to the gap between RL and the untrained base ($0.257$). 
Furthermore, the best single-judge RaR
($0.339$) comfortably exceeds Qwen3-4B ($0.307$), while the worst single-judge RaR ($0.296$) does worse than Qwen3-4B.
A similar sensitivity to judge
choice is observed on GPQA Diamond (Figure~\ref{fig:gpqa-bar}), where the single-judge spread is
$0.345$--$0.406$. Judge identity therefore
has large effects on RaR outcomes under fixed rubric, with spread
comparable to the gap between RL and the untrained base.
This motivates treating judge selection as a key design
variable.

\paragraph{Per-theme breakdown.} We break down performance 
of selected judges by
HealthBench theme (Figure~\ref{fig:per-theme-ensemble}, showing each judge's
delta in mean reward vs.\ the Qwen3-4B-Base baseline) and find that no single judge dominates;
different judges have complementary
strengths across themes. The full per-category breakdown is in
Appendix~\ref{app:per-category}.

\paragraph{Length hacking.}
Reward models and LLM judges have been shown to favor longer
responses \citep{singhal2023long, saito2023verbosity, chen2024odin}. In our experiments, Table~\ref{tab:length} shows that
RaR training under several judges inflates response length
dramatically. 97\% of
gpt-oss-120b's post-training responses hit a 4k-character ($\sim$1024-token) cap vs.\ $\sim$4\% for the
untrained Qwen3-4B-Base. We explore two interventions that partially recover this: (i) an explicit length penalty during
training (\texttt{4o-mini\_lenpen}), which keeps responses short with
no reward loss; (ii) LoRA fine-tuning~\citep{hu2022lora}, which we find empirically dampens the
length expansion (mini-LoRA: $0.295$ reward, $1{,}058$ mean characters vs.\
full-FT mini: $0.296$ reward, $2{,}930$ mean characters). The LoRA dampening
generalizes to RaRE:
\texttt{Gem+DS} (mean+drop) 
scores
$0.330$ under full-FT vs.\ $0.295$ under LoRA, a $\sim$3\,pp cost
for substantially shorter outputs and lower training compute
(Appendix~\ref{app:lora-table}). We use full-FT for headline numbers,
but the gap is small enough that LoRA remains a useful tradeoff.

\begin{table}[t]
\centering
\small
\begin{tabular}{lccc}
\toprule
Judge & Score & Chars & \% $\ge$ 4k \\
\midrule
\multicolumn{4}{l}{\emph{No-RL baselines}} \\
Qwen3-4B-Base      & 0.266          & 1{,}610 &  4.2\% \\
Qwen3-4B           & 0.308          & 1{,}340 &  0.5\% \\
\midrule
\multicolumn{4}{l}{\emph{Single-judge RaR}} \\
gpt-oss-120b       & 0.208          & 4{,}742 & 97.0\% \\
gpt-4o-mini        & 0.296          & 3{,}869 & 42.9\% \\
gemini-2.5-flash-lite & 0.310       & 3{,}016 & 16.3\% \\
deepseek-v3.1      & \textbf{0.317} & 4{,}260 & 76.4\% \\
\midrule
\multicolumn{4}{l}{\emph{Length-hacking interventions}} \\
4o-mini + LoRA     & 0.295          & 1{,}629 &  1.4\% \\
4o-mini + lenpen   & 0.314 & 1{,}815 &  0.2\% \\
\bottomrule
\end{tabular}
\caption{Reward vs.\ response-length statistics on HealthBench 1K subset by training judge.
\emph{Chars} is the mean response length in characters; \emph{\% $\ge$
4k} is the fraction of responses that hit the 4{,}000-character
truncation cap. Single-judge RaR uses full-parameter fine-tuning by
default; the bottom block shows two interventions on the gpt-4o-mini
judge that mitigate length hacking without sacrificing reward.}
\label{tab:length}
\end{table}

\subsection{RaRE Outperforms Single-Judge RaR}
\label{sec:naive-ensemble}

Based on the per-theme complementarity in
Figure~\ref{fig:per-theme-ensemble}, we instantiate RaRE with two judges.
We use two judge pairs in our experiments: Gemini-2.5-flash-lite
with DeepSeek-V3.1 (\texttt{Gem+DS}) for hard-Likert configurations,
and DeepSeek-V3.1 with GPT-4o-mini (\texttt{DS+Mini}) for soft-Likert
configurations, since both endpoints of the latter pair expose the
token-level logprobs that soft-Likert requires. Each pair was chosen
so that the two judges differ in vendor, scale, and training data.

We evaluate the three ensemble strategies from \S\ref{sec:methods}
(naive mean, soft-Likert, drop-on-disagreement) and their
composition. We sweep RaRE's drop-on-disagreement parameter $\tau$ and refer to the best settings per ensemble and dataset simply as \textit{drop} (see Tables~\ref{tab:tau-sweep} and \ref{tab:tau-sweep-gpqa}).
Figures~\ref{fig:healthbench-bar} and \ref{fig:gpqa-bar}
report ensemble results on HealthBench and GPQA Diamond.

The HealthBench score from ensembling the naive mean of Gemini and DeepSeek's integer ratings (0.307) sits between the two constituent judges (0.296 Gemini, 0.338 DeepSeek). Naive averaging alone does not outperform single judges; the gains we report from RaRE below come from filtering or reweighting, not averaging.

Adding drop-on-disagreement (mean + drop) to the Gemini+DeepSeek naive mean (0.347) lifts the score
$+0.8$\,pp above the best single judge by filtering the
$\sim$5\% of rollouts where the two judges disagree most. Soft-Likert
on DeepSeek+GPT-4o-mini without drop (0.330) slightly \emph{hurts}
performance compared to the constituent judges: 
low-confidence disagreement rollouts get down-weighted
but not removed, and their residual contribution may distort the GRPO
advantage estimate. Composing soft-Likert with \textit{drop} on the same
pair ($\mathbf{0.364}$) zeroes the high-disagreement rollouts and
lets the soft-Likert reward on the rest become the strongest signal
we have, exceeding the best single judge by $+2.5$\,pp. The two
mechanisms therefore target different uncertainty sources and compose. 

On GPQA Diamond, we see that ensembling Gemini+DeepSeek naive mean (0.363) also sits close to the two constituent judges (0.360 Gemini, 0.345 DeepSeek). However, with a stronger pair of judges, DeepSeek+4o-mini (0.400 4o-mini), the soft-Likert ensemble performs better (0.372), and adding drop-on-disagreement lifts score to 0.412, exceeding the best single judge.

\begin{figure}[t]
\centering
\begin{tikzpicture}[
  font=\footnotesize,
  x=11.5mm, y=6mm,
  cellv/.style={font=\scriptsize, anchor=center},
  colhdr/.style={font=\scriptsize, align=center, anchor=south, inner sep=2pt},
  rowhdr/.style={font=\scriptsize, anchor=east, inner sep=2pt},
]
\foreach \jname/\jcol in {%
  {DeepSeek\\V3.1}/0,
  {Gemini-2.5\\flash-lite}/1,
  {GPT-4o\\mini}/2,
  {DeepSeek\\+Gemini}/3,
  {DeepSeek+\\GPT-4o-mini}/4}
  \node[colhdr] at (\jcol+0.5, 0.1) {\jname};
\foreach \theme/\row in {%
  Comm./0,
  Complex/1,
  Context/2,
  Emerg./3,
  Global/4,
  Data/5,
  Hedging/6}{
  \node[rowhdr] at (-0.05, -\row-0.5) {\theme};
}
\fill[blue!40] (0,-1) rectangle (1, 0); \node[cellv] at (0.5,-0.5) {$+0.086$};
\fill[blue!28] (1,-1) rectangle (2, 0); \node[cellv] at (1.5,-0.5) {$+0.045$};
\fill[blue!40] (2,-1) rectangle (3, 0); \node[cellv] at (2.5,-0.5) {$+0.082$};
\fill[blue!40] (3,-1) rectangle (4, 0); \node[cellv] at (3.5,-0.5) {$+0.090$};
\fill[blue!40] (4,-1) rectangle (5, 0); \node[cellv] at (4.5,-0.5) {$\mathbf{+0.099}$};
\fill[red!15]  (0,-2) rectangle (1,-1); \node[cellv] at (0.5,-1.5) {$-0.024$};
\fill[red!15]  (1,-2) rectangle (2,-1); \node[cellv] at (1.5,-1.5) {$-0.029$};
\fill[blue!15] (2,-2) rectangle (3,-1); \node[cellv] at (2.5,-1.5) {$+0.026$};
\fill[white]   (3,-2) rectangle (4,-1); \node[cellv] at (3.5,-1.5) {$-0.001$};
\fill[blue!15] (4,-2) rectangle (5,-1); \node[cellv] at (4.5,-1.5) {$\mathbf{+0.039}$};
\fill[blue!50] (0,-3) rectangle (1,-2); \node[cellv] at (0.5,-2.5) {$+0.111$};
\fill[blue!28] (1,-3) rectangle (2,-2); \node[cellv] at (1.5,-2.5) {$+0.051$};
\fill[blue!40] (2,-3) rectangle (3,-2); \node[cellv] at (2.5,-2.5) {$+0.098$};
\fill[blue!50] (3,-3) rectangle (4,-2); \node[cellv] at (3.5,-2.5) {$+0.112$};
\fill[blue!50] (4,-3) rectangle (5,-2); \node[cellv] at (4.5,-2.5) {$\mathbf{+0.118}$};
\fill[blue!40] (0,-4) rectangle (1,-3); \node[cellv] at (0.5,-3.5) {$+0.075$};
\fill[blue!15] (1,-4) rectangle (2,-3); \node[cellv] at (1.5,-3.5) {$+0.023$};
\fill[blue!28] (2,-4) rectangle (3,-3); \node[cellv] at (2.5,-3.5) {$+0.060$};
\fill[blue!40] (3,-4) rectangle (4,-3); \node[cellv] at (3.5,-3.5) {$+0.072$};
\fill[blue!40] (4,-4) rectangle (5,-3); \node[cellv] at (4.5,-3.5) {$\mathbf{+0.089}$};
\fill[blue!50] (0,-5) rectangle (1,-4); \node[cellv] at (0.5,-4.5) {$+0.102$};
\fill[blue!28] (1,-5) rectangle (2,-4); \node[cellv] at (1.5,-4.5) {$+0.041$};
\fill[blue!40] (2,-5) rectangle (3,-4); \node[cellv] at (2.5,-4.5) {$+0.097$};
\fill[blue!50] (3,-5) rectangle (4,-4); \node[cellv] at (3.5,-4.5) {$+0.106$};
\fill[blue!50] (4,-5) rectangle (5,-4); \node[cellv] at (4.5,-4.5) {$\mathbf{+0.131}$};
\fill[blue!28] (0,-6) rectangle (1,-5); \node[cellv] at (0.5,-5.5) {$+0.069$};
\fill[blue!15] (1,-6) rectangle (2,-5); \node[cellv] at (1.5,-5.5) {$+0.034$};
\fill[blue!40] (2,-6) rectangle (3,-5); \node[cellv] at (2.5,-5.5) {$+0.077$};
\fill[blue!40] (3,-6) rectangle (4,-5); \node[cellv] at (3.5,-5.5) {$+0.080$};
\fill[blue!40] (4,-6) rectangle (5,-5); \node[cellv] at (4.5,-5.5) {$\mathbf{+0.088}$};
\fill[blue!40] (0,-7) rectangle (1,-6); \node[cellv] at (0.5,-6.5) {$+0.093$};
\fill[blue!28] (1,-7) rectangle (2,-6); \node[cellv] at (1.5,-6.5) {$+0.065$};
\fill[blue!40] (2,-7) rectangle (3,-6); \node[cellv] at (2.5,-6.5) {$+0.098$};
\fill[blue!50] (3,-7) rectangle (4,-6); \node[cellv] at (3.5,-6.5) {$+0.114$};
\fill[blue!50] (4,-7) rectangle (5,-6); \node[cellv] at (4.5,-6.5) {$\mathbf{+0.134}$};
\draw[black!30, thin, step=1] (0,0) grid (5,-7);
\begin{scope}[shift={(0,-7.8)}]
  \fill[red!15]  (0.5,0) rectangle ++(0.5,0.35);
  \fill[white]   (1.0,0) rectangle ++(0.5,0.35);
  \fill[blue!15] (1.5,0) rectangle ++(0.5,0.35);
  \fill[blue!28] (2.0,0) rectangle ++(0.5,0.35);
  \fill[blue!40] (2.5,0) rectangle ++(0.5,0.35);
  \fill[blue!50] (3.0,0) rectangle ++(0.5,0.35);
  \draw[black!30, thin] (0.5,0) rectangle (3.5, 0.35);
  \node[font=\scriptsize, anchor=east] at (0.45, 0.175) {worse};
  \node[font=\scriptsize, anchor=west] at (3.55, 0.175) {better};
\end{scope}
\end{tikzpicture}
\caption{Mean-reward delta on English-only HealthBench (n=3931)
vs.\ Qwen3-4B-Base across themes (Comm.\ = communication;
Emerg.\ = emergency referrals). Columns: single judges
(DeepSeek-V3.1, Gemini-2.5-flash-lite, GPT-4o-mini) and ensembles
(DeepSeek+Gemini, naive mean; DeepSeek+GPT-4o-mini, soft-Likert; both
with drop). Bold: row winner.}
\label{fig:per-theme-ensemble}
\end{figure}

\paragraph{Complementary judge strengths.}
Figure~\ref{fig:per-theme-ensemble} shows that single judges win on disjoint
themes on HealthBench: DeepSeek wins \texttt{communication},
\texttt{context\_seeking}, \texttt{emergency\_referrals},
and \texttt{global\_health}; GPT-4o-mini wins
\texttt{complex\_responses}, \texttt{data\_tasks}, and 
\texttt{hedging}. Furthermore, our best RaRE configuration (soft +
drop on DeepSeek+GPT-4o-mini) combines the strengths of its two
constituent judges and achieves the best results in each theme
across the board. RaRE thus inherits, rather than averages out, the
constituent judges' theme-level strengths. In Appendix Figure~\ref{fig:per-topic-gpqa}, we similarly report per-topic accuracy on GPQA Diamond, which demonstrates the qualities of different judges.

\subsection{Configuration Ablations}
\label{sec:disagreement-ensemble}

Our strongest configuration combines soft-Likert with a disagreement
threshold of $\tau{=}0.30$ on medicine and $\tau{=}0.40$ on science. Here we show that each piece of this
configuration is important: changing $\tau$, removing the
soft-Likert reward, or using alternative confidence weighting schemes all reduce the score substantially.

\paragraph{Threshold sweep ($\tau$).}
We sweep $\tau$ on the Gemini+DeepSeek pair under the hard-Likert
reward path and find $\tau{=}0.30$ is best on the HealthBench 1K subset (Table~\ref{tab:tau-sweep}),
with $\sim$1.6\,pp lift over no drop and scores dropping on either
side. $\tau{=}0.15$ is too aggressive ($\sim$14\% of rollouts masked,
gradient signal too sparse), and $\tau{=}0.35$ is too permissive
($\sim$2\% masked). $\tau{=}0.30$ corresponds to a mean-rating gap of $\sim$1.2 on
the $[1,5]$ scale: the policy benefits most from filtering rollouts
where the judges fundamentally disagree, not where they differ by a
small amount. Interestingly, this threshold also transfers well to the DeepSeek+4o-mini soft-Likert pair, which is our best result (Appendix \ref{app:dsmini-tau-sweep}).

Similarly, we found $\tau=0.40$
to be the best disagreement threshold for the DeepSeek+4o-mini pair under the soft-Likert reward path on GPQA
by sweeping through $\tau$ and selecting the result with the best score on the validation split, with results reported in Table \ref{tab:tau-sweep-gpqa}. We find
that $\tau=0.40$ is the sweet spot, filtering $\sim$24\% of 
rollouts. Interestingly, RaR-Science requires filtering
a greater percentage of rollouts than medicine, suggesting that
there is inherently much more judge disagreement on what a good
science response should be.

\begin{table}[!ht]
\centering
\small
\setlength{\tabcolsep}{4pt}
\begin{tabular}{@{}lrcc@{}}
\toprule
$\tau$ & drop\,\% & Score & 95\% CI \\
\midrule
none (mean only)     & 0\%   & $0.314$          & {\footnotesize $[0.295,\,0.335]$} \\
$0.15$               & 14\%  & $0.251$          & {\footnotesize $[0.233,\,0.271]$} \\
$0.25$               & 9\%   & $0.301$          & {\footnotesize $[0.281,\,0.323]$} \\
$\mathbf{0.30}$      & 5\%   & $\mathbf{0.330}$ & {\footnotesize $[0.312,\,0.351]$} \\
$0.35$               & 2\%   & $0.273$          & {\footnotesize $[0.254,\,0.294]$} \\
\bottomrule
\end{tabular}
\caption{Drop-on-disagreement threshold sweep on the Gemini+DeepSeek
pair (hard-Likert reward path) on the HealthBench 1K subset. drop\,\% is the mean fraction of
rollouts masked across training steps; $\tau{=}0.30$ achieves the
best score. 95\% CIs computed via bootstrap.}
\label{tab:tau-sweep}
\end{table}

\begin{table}[!ht]
\centering
\small
\begin{tabular}{lrc}
\toprule
$\tau$ & drop\,\% & Validation\ score \\
\midrule
none (mean only)     & 0\%   & $0.349$     \\
$0.30$               & 29\%  & $0.362$     \\
$\mathbf{0.40}$      & 24\%  & $\mathbf{0.380}$ \\
$0.45$               & 25\%  & $0.356$     \\
$0.50$               & 12\%  & $0.374$     \\
\bottomrule
\end{tabular}
\caption{Drop-on-disagreement threshold sweep on the DeepSeek+GPT-4o-mini
pair (soft-Likert reward) on GPQA Diamond. Scores are 5-seed means on a
50\% validation split; drop\,\% is
the mean fraction of rollouts masked across training steps.
$\tau{=}0.40$ is selected as the best validation threshold and is evaluated on the disjoint 50\% test split in the main results.}
\label{tab:tau-sweep-gpqa}
\end{table}

\paragraph{Soft-Likert vs.\ hard-Likert.}
On the DeepSeek+GPT-4o-mini pair with \textit{drop} fixed, switching from
hard-Likert (integer ratings) to soft-Likert (inverse-variance
confidence weighting) lifts the score from $0.295$ to
$\mathbf{0.364}$ ($+6.9$\,pp). Inverse-variance confidence
weighting (soft-Likert) therefore adds large lift on top of the drop mask.

\paragraph{Confidence-aggregation ablation.}
We ablate inverse-variance confidence (Eq.~\ref{eq:conf}) with two aggregation schemes that also derive a per-judge weight from
$p_j^{(i)}$. \textbf{Top-logit} uses the modal mass
$c_j^{(i)} = \max_k p_j^{(i)}(k)$, which ignores how the
distribution is spread.
\textbf{Winner-takes-all} drops weighted averaging entirely and uses
the rating of the single judge with the highest top-logit confidence.

Both alternatives collapse (Table~\ref{tab:conf-ablation}): top-logit
scores $0.214$ ($-12$\,pp), winner-takes-all
scores $0.198$ ($-14$\,pp), both below the untrained base, showing that confidence aggregation must account for the distribution of the judge's confidence.

\begin{table}[t]
\centering
\small
\setlength{\tabcolsep}{4pt}
\begin{tabular}{@{}lcc@{}}
\toprule
Confidence + aggregation & Score & 95\% CI \\
\midrule
inverse-var, weighted avg        & $\mathbf{0.336}$ & {\footnotesize $[0.316,\,0.355]$} \\
top-logit, weighted avg          & $0.214$          & {\footnotesize $[0.195,\,0.233]$} \\
top-logit, winner-takes-all      & $0.198$          & {\footnotesize $[0.179,\,0.217]$} \\
\bottomrule
\end{tabular}
\caption{Confidence-aggregation ablation on the
DeepSeek+GPT-4o-mini pair with drop030, evaluated on the HealthBench 1K subset. Replacing inverse-variance
(Eq.~\ref{eq:conf}) with simpler schemes drops the score by 12--14\,pp
and below the untrained base ($0.266$). 95\% CIs computed via
bootstrap.}
\label{tab:conf-ablation}
\end{table}

\section{Discussion}

\paragraph{Judge choice matters in RaR.} Varying the judge under a fixed rubric produces a $\sim$4.3\,pp spread in HealthBench score across five judges. This spread is greater than the gap between Qwen3-4B-Base and the lowest single-judge RaR run. Prior RaR work has focused almost exclusively on improving the rubric, treating the judge as fixed. Our results suggest that the judge side carries comparable signal and deserves comparable attention.

\paragraph{Ensembles win by combining complementary judges, not by averaging.} We might expect a judge ensemble to smooth out variance from any single judge, with the ensemble converging toward the strongest judge. We do not find this. RaRE consistently improves on HealthBench themes beyond either of its two judges individually (Figure~\ref{fig:per-theme-ensemble}), suggesting that the gain comes from combining the judges' non-overlapping strengths, rather than from selecting the stronger judge. Naive averaging does not extract this gain: the mean of judges' ratings averages their individual scores and does not outperform the best single judge. The gain only appears once we filter high-disagreement rollouts.

\paragraph{Disagreement filtering is the most impactful mechanism.} Across our ablations, the drop-on-disagreement mask contributes the most. Adding it to the naive Gemini+DeepSeek mean ensemble lifts the score by $+4.0$\,pp (0.307 to 0.347); removing it from the DeepSeek+4o-mini soft+drop configuration costs $3.4$\,pp (0.364 to 0.330). The soft-Likert reward provides additional gain when composed with the mask, but on its own does not outperform the best single judge on HealthBench. The broader LLM-as-a-verifier literature \citep{lu2025does, singh2026v_1, kwok2026llmverifier} reports a similar finding in the evaluation setting, where explicit disagreement filtering matters more than the choice of any single judge. Our results extend this from evaluation to training.

\paragraph{Relation to concurrent and complementary work.} Two recent directions are complementary to ours. Online rubric updates \citep{rezaei2025online, li2026evolm} reduce reward-signal variance by improving rubric content during training; we reduce it by improving judge aggregation. The two operate on different parts of the RaR pipeline and should compose, though we leave that combination to future work. Concurrent work by \citet{mahmoud2026rewardhackingrubricbasedreinforcement} also uses multiple judges in the RaR setting, but in the opposite direction: they train against a single verifier and use a cross-family panel of three frontier judges to \emph{diagnose} reward hacking, while we use multiple judges to \emph{produce} the training signal. The two contributions are independent and could be combined, with RaRE serving as the training-time signal and their panel as the evaluation-time check.

\section{Conclusion}

We proposed RaRE (Rubrics as Rewards with Ensembles), combining a soft-Likert confidence-aggregation reward with a drop-on-disagreement mask, which outperforms every single-judge RaR variant on both HealthBench and GPQA-Diamond. Our results suggest RaRE's gains arise from combining complementary judge strengths through confidence and disagreement aggregation.

\section*{Limitations}
Due to limited compute, our experiments are restricted to a single base policy (Qwen3-4B-Base) evaluated on two domains: HealthBench (medicine) and GPQA Diamond (science), so our findings may not transfer to other policy families,
model scales, or other domains. We mitigate this
with per-theme and per-category breakdowns and extensive ablations,
but broader validation is left to future work.

We evaluate HealthBench with GPT-4o-mini, following the convention established by \citet{arora2025healthbench} and \citet{gunjal2025rubrics}, which may introduce evaluation-judge bias since it is also used as a training judge. To test whether this overlap affects our results, we regrade policy responses with GLM-5.3-Flash, a judge from a different model family (Appendix~\ref{app:alt-evaluator}). Relative rankings remain similar, and RaRE ranks first under both judges. A larger multi-judge evaluation panel, such as the one used by \citet{mahmoud2026rewardhackingrubricbasedreinforcement}, would provide additional robustness against evaluation-judge bias; our GPQA Diamond results, which use verifier-style evaluation rather than an LLM judge, partially address this concern.

RaRE's improvement on GPQA Diamond compared to single-judge RaR is not statistically significant across 5 training seeds. Running additional seeds could help tighten confidence intervals and allow for a more definitive comparison, which we leave to future work due to compute constraints.

When applying RaRE's drop-on-disagreement mask, we discard the dropped rollouts entirely rather than extracting any partial signal from them. Initial exploratory attempts in this direction (Appendix~\ref{app:retro}) did not yield any improvement over our baseline methods, but further exploration is left to future work.

Finally, RaRE depends on inter-judge disagreement as a
signal. When judges largely agree, RaRE provides little
additional signal beyond a single judge; a more diverse judge pool
could expand the regime where ensembling helps.

\section*{Ethical Considerations}

Our work studies how judge selection and judge ensembling affect
rubric-based RL training, with the goal of making the reward signal
more reliable. Because the resulting policies are trained against
LLM-judge rewards, they inherit biases of the judges used: RaRE
reduces idiosyncratic per-judge bias but does not
eliminate biases shared across judges, and the drop-on-disagreement
filter removes precisely those rollouts where judges disagree most,
which could exclude cases where a single judge correctly flags a
safety-relevant issue that others miss. We therefore encourage
practitioners who deploy ensemble-trained policies, particularly in
health or scientific contexts, to audit them separately for these
biases rather than treat ensembling as a substitute for evaluation,
and we treat the models in this work as research artifacts rather than
clinical or scientific advisors.

\section*{Acknowledgments}
We would like to thank Harman Singh for feedback on this work and helpful discussions. AH was supported by the National Defense Science and Engineering Graduate (NDSEG) Fellowship Program. This material is based upon work supported by the Air Force Office of Scientific Research under award number FA9550-25-C-B010 in the amount of \$143,000. YB is supported by ONR MURI award N00014-21-1-2801.

\bibliography{custom}

\clearpage

\appendix

\section{Full per-category breakdown}
\label{app:per-category}

Figure \ref{fig:per-category} and Figure \ref{fig:per-topic-gpqa} give the per-judge delta vs.\
Qwen3-4B-Base on every HealthBench category and GPQA Diamond topic, respectively, complementing the
per-theme summary in Figure \ref{fig:per-theme-ensemble}. \textbf{Bold} marks
the per-category/topic winner. Our best RaRE configuration (\texttt{dsmini}) wins 12 of 17
HealthBench categories outright. It also wins 1 GPQA topic (biology) outright, with strong performances on the other 2 topics.

\begin{figure*}[t]
\centering
\begin{tikzpicture}[
  font=\small,
  x=13mm, y=7mm,
  cellv/.style={font=\small\sffamily, anchor=center},
  colhdr/.style={font=\bfseries, anchor=base, inner sep=2pt},
  rowhdr/.style={font=\ttfamily\scriptsize, anchor=east, inner sep=3pt},
  rowmeta/.style={font=\scriptsize, anchor=west,
                  inner sep=3pt, text=black!55},
]
\foreach \jname/\jcol in {ds/0, gem/1, mini/2, qw/3, oss/4, gemds/5, dsmini/6}
  \node[colhdr] at (\jcol+0.5, 0.25) {\jname};

\foreach \cat/\nbase/\row in {%
  any-reducible-uncertainty/{n=242,\ base 0.216}/0,
  conditionally-emergent/{n=168,\ base 0.290}/1,
  context-does-not-matter/{n=142,\ base 0.274}/2,
  context-matters-but-unclear/{n=148,\ base 0.155}/3,
  context-matters-is-clear/{n=146,\ base 0.204}/4,
  detailed/{n=122,\ base 0.170}/5,
  emergent/{n=132,\ base 0.381}/6,
  enough-context/{n=207,\ base 0.295}/7,
  enough-info-to-complete-task/{n=214,\ base 0.345}/8,
  health-professional/{n=297,\ base 0.432}/9,
  no-uncertainty/{n=204,\ base 0.301}/10,
  non-emergent/{n=123,\ base 0.505}/11,
  not-enough-context/{n=158,\ base 0.103}/12,
  not-enough-info-to-complete-task/{n=168,\ base 0.092}/13,
  not-health-professional/{n=350,\ base 0.450}/14,
  only-irreducible-uncertainty/{n=182,\ base 0.280}/15,
  simple/{n=146,\ base 0.209}/16%
}{
  \node[rowhdr]  at (-0.05, -\row-0.5) {\cat};
  \node[rowmeta] at ( 7.05, -\row-0.5) {\nbase};
}

\fill[blue!40] (0,-1) rectangle (1, 0); \node[cellv] at (0.5,-0.5) {$+0.090$};
\fill[blue!28] (1,-1) rectangle (2, 0); \node[cellv] at (1.5,-0.5) {$+0.059$};
\fill[blue!40] (2,-1) rectangle (3, 0); \node[cellv] at (2.5,-0.5) {$+0.091$};
\fill[blue!28] (3,-1) rectangle (4, 0); \node[cellv] at (3.5,-0.5) {$+0.053$};
\fill[blue!28] (4,-1) rectangle (5, 0); \node[cellv] at (4.5,-0.5) {$+0.068$};
\fill[blue!40] (5,-1) rectangle (6, 0); \node[cellv] at (5.5,-0.5) {$+0.109$};
\fill[blue!40] (6,-1) rectangle (7, 0); \node[cellv] at (6.5,-0.5) {$\mathbf{+0.120}$};
\fill[blue!40] (0,-2) rectangle (1,-1); \node[cellv] at (0.5,-1.5) {$+0.175$};
\fill[blue!28] (1,-2) rectangle (2,-1); \node[cellv] at (1.5,-1.5) {$+0.069$};
\fill[blue!40] (2,-2) rectangle (3,-1); \node[cellv] at (2.5,-1.5) {$+0.151$};
\fill[blue!40] (3,-2) rectangle (4,-1); \node[cellv] at (3.5,-1.5) {$+0.125$};
\fill[blue!40] (4,-2) rectangle (5,-1); \node[cellv] at (4.5,-1.5) {$+0.095$};
\fill[blue!40] (5,-2) rectangle (6,-1); \node[cellv] at (5.5,-1.5) {$+0.170$};
\fill[blue!40] (6,-2) rectangle (7,-1); \node[cellv] at (6.5,-1.5) {$\mathbf{+0.182}$};
\fill[blue!40] (0,-3) rectangle (1,-2); \node[cellv] at (0.5,-2.5) {$+0.123$};
\fill[blue!15] (1,-3) rectangle (2,-2); \node[cellv] at (1.5,-2.5) {$+0.035$};
\fill[blue!40] (2,-3) rectangle (3,-2); \node[cellv] at (2.5,-2.5) {$+0.073$};
\fill[blue!28] (3,-3) rectangle (4,-2); \node[cellv] at (3.5,-2.5) {$+0.068$};
\fill[blue!28] (4,-3) rectangle (5,-2); \node[cellv] at (4.5,-2.5) {$+0.066$};
\fill[blue!40] (5,-3) rectangle (6,-2); \node[cellv] at (5.5,-2.5) {$+0.125$};
\fill[blue!40] (6,-3) rectangle (7,-2); \node[cellv] at (6.5,-2.5) {$\mathbf{+0.132}$};
\fill[blue!40] (0,-4) rectangle (1,-3); \node[cellv] at (0.5,-3.5) {$+0.080$};
\fill[blue!15] (1,-4) rectangle (2,-3); \node[cellv] at (1.5,-3.5) {$+0.048$};
\fill[blue!40] (2,-4) rectangle (3,-3); \node[cellv] at (2.5,-3.5) {$+0.085$};
\fill[blue!28] (3,-4) rectangle (4,-3); \node[cellv] at (3.5,-3.5) {$+0.051$};
\fill[blue!28] (4,-4) rectangle (5,-3); \node[cellv] at (4.5,-3.5) {$+0.068$};
\fill[blue!40] (5,-4) rectangle (6,-3); \node[cellv] at (5.5,-3.5) {$+0.091$};
\fill[blue!40] (6,-4) rectangle (7,-3); \node[cellv] at (6.5,-3.5) {$\mathbf{+0.134}$};
\fill[blue!40] (0,-5) rectangle (1,-4); \node[cellv] at (0.5,-4.5) {$+0.076$};
\fill[blue!15] (1,-5) rectangle (2,-4); \node[cellv] at (1.5,-4.5) {$+0.027$};
\fill[blue!40] (2,-5) rectangle (3,-4); \node[cellv] at (2.5,-4.5) {$+0.092$};
\fill[blue!28] (3,-5) rectangle (4,-4); \node[cellv] at (3.5,-4.5) {$+0.066$};
\fill[blue!40] (4,-5) rectangle (5,-4); \node[cellv] at (4.5,-4.5) {$+0.087$};
\fill[blue!40] (5,-5) rectangle (6,-4); \node[cellv] at (5.5,-4.5) {$+0.079$};
\fill[blue!40] (6,-5) rectangle (7,-4); \node[cellv] at (6.5,-4.5) {$\mathbf{+0.119}$};
\fill[white]   (0,-6) rectangle (1,-5); \node[cellv] at (0.5,-5.5) {$-0.003$};
\fill[red!15]  (1,-6) rectangle (2,-5); \node[cellv] at (1.5,-5.5) {$-0.024$};
\fill[blue!15] (2,-6) rectangle (3,-5); \node[cellv] at (2.5,-5.5) {$+0.044$};
\fill[white]   (3,-6) rectangle (4,-5); \node[cellv] at (3.5,-5.5) {$+0.015$};
\fill[blue!28] (4,-6) rectangle (5,-5); \node[cellv] at (4.5,-5.5) {$\mathbf{+0.056}$};
\fill[blue!15] (5,-6) rectangle (6,-5); \node[cellv] at (5.5,-5.5) {$+0.021$};
\fill[blue!15] (6,-6) rectangle (7,-5); \node[cellv] at (6.5,-5.5) {$+0.025$};
\fill[white]   (0,-7) rectangle (1,-6); \node[cellv] at (0.5,-6.5) {$\mathbf{-0.011}$};
\fill[red!28]  (1,-7) rectangle (2,-6); \node[cellv] at (1.5,-6.5) {$-0.063$};
\fill[red!15]  (2,-7) rectangle (3,-6); \node[cellv] at (2.5,-6.5) {$-0.049$};
\fill[red!15]  (3,-7) rectangle (4,-6); \node[cellv] at (3.5,-6.5) {$-0.036$};
\fill[red!15]  (4,-7) rectangle (5,-6); \node[cellv] at (4.5,-6.5) {$-0.026$};
\fill[red!15]  (5,-7) rectangle (6,-6); \node[cellv] at (5.5,-6.5) {$-0.028$};
\fill[red!15]  (6,-7) rectangle (7,-6); \node[cellv] at (6.5,-6.5) {$-0.034$};
\fill[blue!40] (0,-8) rectangle (1,-7); \node[cellv] at (0.5,-7.5) {$+0.101$};
\fill[blue!15] (1,-8) rectangle (2,-7); \node[cellv] at (1.5,-7.5) {$+0.039$};
\fill[blue!40] (2,-8) rectangle (3,-7); \node[cellv] at (2.5,-7.5) {$+0.087$};
\fill[blue!40] (3,-8) rectangle (4,-7); \node[cellv] at (3.5,-7.5) {$+0.071$};
\fill[blue!28] (4,-8) rectangle (5,-7); \node[cellv] at (4.5,-7.5) {$+0.054$};
\fill[blue!40] (5,-8) rectangle (6,-7); \node[cellv] at (5.5,-7.5) {$+0.103$};
\fill[blue!40] (6,-8) rectangle (7,-7); \node[cellv] at (6.5,-7.5) {$\mathbf{+0.113}$};
\fill[blue!28] (0,-9) rectangle (1,-8); \node[cellv] at (0.5,-8.5) {$+0.068$};
\fill[blue!28] (1,-9) rectangle (2,-8); \node[cellv] at (1.5,-8.5) {$+0.050$};
\fill[blue!40] (2,-9) rectangle (3,-8); \node[cellv] at (2.5,-8.5) {$+0.075$};
\fill[blue!40] (3,-9) rectangle (4,-8); \node[cellv] at (3.5,-8.5) {$+0.074$};
\fill[blue!40] (4,-9) rectangle (5,-8); \node[cellv] at (4.5,-8.5) {$\mathbf{+0.106}$};
\fill[blue!40] (5,-9) rectangle (6,-8); \node[cellv] at (5.5,-8.5) {$+0.076$};
\fill[blue!40] (6,-9) rectangle (7,-8); \node[cellv] at (6.5,-8.5) {$+0.096$};
\fill[blue!40] (0,-10) rectangle (1,-9); \node[cellv] at (0.5,-9.5) {$+0.073$};
\fill[blue!15] (1,-10) rectangle (2,-9); \node[cellv] at (1.5,-9.5) {$+0.048$};
\fill[blue!40] (2,-10) rectangle (3,-9); \node[cellv] at (2.5,-9.5) {$\mathbf{+0.079}$};
\fill[blue!28] (3,-10) rectangle (4,-9); \node[cellv] at (3.5,-9.5) {$+0.064$};
\fill[blue!40] (4,-10) rectangle (5,-9); \node[cellv] at (4.5,-9.5) {$+0.074$};
\fill[blue!40] (5,-10) rectangle (6,-9); \node[cellv] at (5.5,-9.5) {$+0.070$};
\fill[blue!40] (6,-10) rectangle (7,-9); \node[cellv] at (6.5,-9.5) {$+0.070$};
\fill[blue!40] (0,-11) rectangle (1,-10); \node[cellv] at (0.5,-10.5) {$+0.102$};
\fill[blue!28] (1,-11) rectangle (2,-10); \node[cellv] at (1.5,-10.5) {$+0.061$};
\fill[blue!40] (2,-11) rectangle (3,-10); \node[cellv] at (2.5,-10.5) {$+0.081$};
\fill[blue!40] (3,-11) rectangle (4,-10); \node[cellv] at (3.5,-10.5) {$+0.078$};
\fill[blue!40] (4,-11) rectangle (5,-10); \node[cellv] at (4.5,-10.5) {$+0.098$};
\fill[blue!40] (5,-11) rectangle (6,-10); \node[cellv] at (5.5,-10.5) {$+0.118$};
\fill[blue!40] (6,-11) rectangle (7,-10); \node[cellv] at (6.5,-10.5) {$\mathbf{+0.126}$};
\fill[white]   (0,-12) rectangle (1,-11); \node[cellv] at (0.5,-11.5) {$+0.015$};
\fill[blue!15] (1,-12) rectangle (2,-11); \node[cellv] at (1.5,-11.5) {$+0.042$};
\fill[blue!15] (2,-12) rectangle (3,-11); \node[cellv] at (2.5,-11.5) {$+0.049$};
\fill[white]   (3,-12) rectangle (4,-11); \node[cellv] at (3.5,-11.5) {$-0.014$};
\fill[blue!28] (4,-12) rectangle (5,-11); \node[cellv] at (4.5,-11.5) {$+0.069$};
\fill[blue!15] (5,-12) rectangle (6,-11); \node[cellv] at (5.5,-11.5) {$+0.038$};
\fill[blue!40] (6,-12) rectangle (7,-11); \node[cellv] at (6.5,-11.5) {$\mathbf{+0.093}$};
\fill[blue!40] (0,-13) rectangle (1,-12); \node[cellv] at (0.5,-12.5) {$+0.099$};
\fill[blue!28] (1,-13) rectangle (2,-12); \node[cellv] at (1.5,-12.5) {$+0.056$};
\fill[blue!40] (2,-13) rectangle (3,-12); \node[cellv] at (2.5,-12.5) {$+0.092$};
\fill[blue!40] (3,-13) rectangle (4,-12); \node[cellv] at (3.5,-12.5) {$+0.074$};
\fill[blue!15] (4,-13) rectangle (5,-12); \node[cellv] at (4.5,-12.5) {$+0.038$};
\fill[blue!40] (5,-13) rectangle (6,-12); \node[cellv] at (5.5,-12.5) {$+0.116$};
\fill[blue!40] (6,-13) rectangle (7,-12); \node[cellv] at (6.5,-12.5) {$\mathbf{+0.138}$};
\fill[blue!40] (0,-14) rectangle (1,-13); \node[cellv] at (0.5,-13.5) {$+0.115$};
\fill[blue!15] (1,-14) rectangle (2,-13); \node[cellv] at (1.5,-13.5) {$+0.034$};
\fill[blue!40] (2,-14) rectangle (3,-13); \node[cellv] at (2.5,-13.5) {$\mathbf{+0.123}$};
\fill[blue!28] (3,-14) rectangle (4,-13); \node[cellv] at (3.5,-13.5) {$+0.062$};
\fill[blue!40] (4,-14) rectangle (5,-13); \node[cellv] at (4.5,-13.5) {$+0.121$};
\fill[blue!40] (5,-14) rectangle (6,-13); \node[cellv] at (5.5,-13.5) {$+0.116$};
\fill[blue!40] (6,-14) rectangle (7,-13); \node[cellv] at (6.5,-13.5) {$+0.103$};
\fill[blue!40] (0,-15) rectangle (1,-14); \node[cellv] at (0.5,-14.5) {$+0.101$};
\fill[blue!28] (1,-15) rectangle (2,-14); \node[cellv] at (1.5,-14.5) {$+0.050$};
\fill[blue!40] (2,-15) rectangle (3,-14); \node[cellv] at (2.5,-14.5) {$+0.087$};
\fill[blue!40] (3,-15) rectangle (4,-14); \node[cellv] at (3.5,-14.5) {$+0.079$};
\fill[blue!28] (4,-15) rectangle (5,-14); \node[cellv] at (4.5,-14.5) {$+0.066$};
\fill[blue!40] (5,-15) rectangle (6,-14); \node[cellv] at (5.5,-14.5) {$+0.115$};
\fill[blue!40] (6,-15) rectangle (7,-14); \node[cellv] at (6.5,-14.5) {$\mathbf{+0.124}$};
\fill[blue!40] (0,-16) rectangle (1,-15); \node[cellv] at (0.5,-15.5) {$+0.086$};
\fill[blue!40] (1,-16) rectangle (2,-15); \node[cellv] at (1.5,-15.5) {$+0.074$};
\fill[blue!40] (2,-16) rectangle (3,-15); \node[cellv] at (2.5,-15.5) {$+0.094$};
\fill[white]   (3,-16) rectangle (4,-15); \node[cellv] at (3.5,-15.5) {$+0.002$};
\fill[blue!15] (4,-16) rectangle (5,-15); \node[cellv] at (4.5,-15.5) {$+0.036$};
\fill[blue!40] (5,-16) rectangle (6,-15); \node[cellv] at (5.5,-15.5) {$+0.097$};
\fill[blue!40] (6,-16) rectangle (7,-15); \node[cellv] at (6.5,-15.5) {$\mathbf{+0.139}$};
\fill[red!15]  (0,-17) rectangle (1,-16); \node[cellv] at (0.5,-16.5) {$-0.040$};
\fill[red!15]  (1,-17) rectangle (2,-16); \node[cellv] at (1.5,-16.5) {$-0.034$};
\fill[white]   (2,-17) rectangle (3,-16); \node[cellv] at (2.5,-16.5) {$-0.014$};
\fill[white]   (3,-17) rectangle (4,-16); \node[cellv] at (3.5,-16.5) {$-0.014$};
\fill[white]   (4,-17) rectangle (5,-16); \node[cellv] at (4.5,-16.5) {$+0.008$};
\fill[red!15]  (5,-17) rectangle (6,-16); \node[cellv] at (5.5,-16.5) {$-0.029$};
\fill[blue!15] (6,-17) rectangle (7,-16); \node[cellv] at (6.5,-16.5) {$\mathbf{+0.027}$};

\draw[black!30, thin, step=1] (0,0) grid (7,-17);

\begin{scope}[shift={(0,-18)}]
  \fill[red!40]  (0.0,0) rectangle ++(0.5,0.4);
  \fill[red!28]  (0.5,0) rectangle ++(0.5,0.4);
  \fill[red!15]  (1.0,0) rectangle ++(0.5,0.4);
  \fill[white]   (1.5,0) rectangle ++(0.5,0.4);
  \fill[blue!15] (2.0,0) rectangle ++(0.5,0.4);
  \fill[blue!28] (2.5,0) rectangle ++(0.5,0.4);
  \fill[blue!40] (3.0,0) rectangle ++(0.5,0.4);
  \draw[black!30, thin] (0,0) rectangle (3.5, 0.4);
  \node[font=\scriptsize, anchor=east] at (-0.05, 0.2) {worse $\leftarrow$};
  \node[font=\scriptsize, anchor=west] at ( 3.55, 0.2) {$\rightarrow$ better};
\end{scope}
\end{tikzpicture}
\caption{Per-category delta in English-only HealthBench mean
reward vs.\ Qwen3-4B-Base.
\textbf{Single judges:} \texttt{ds} = DeepSeek-V3.1, \texttt{gem} =
Gemini-2.5-flash-lite, \texttt{mini} = GPT-4o-mini, \texttt{qw} =
Qwen3-235B-Instruct, \texttt{oss} = GPT-OSS-120B.
\textbf{Ensembles:} \texttt{gemds} = naive mean of Gemini+DeepSeek
with drop030; \texttt{dsmini} = soft-Likert of DeepSeek+GPT-4o-mini
with drop030. Diverging red--blue scale; \textbf{bold} marks the
per-category winner. The \texttt{dsmini} ensemble wins 12 of 17
categories outright. \texttt{emergent} is the universal pain point:
all seven RaR variants \emph{worsen} relative to the no-RL base.}
\label{fig:per-category}
\end{figure*}

\begin{figure*}[t]
\centering
\begin{tikzpicture}[
  font=\small,
  x=13mm, y=8mm,
  cellv/.style={font=\small\sffamily, anchor=center},
  colhdr/.style={font=\bfseries, anchor=base, inner sep=2pt},
  rowhdr/.style={font=\ttfamily\scriptsize, anchor=east, inner sep=3pt},
  rowmeta/.style={font=\scriptsize, anchor=west,
                  inner sep=3pt, text=black!55},
]
\foreach \jname/\jcol in {ds/0, gem/1, mini/2, qw/3, oss/4, gemds/5, dsmini/6}
  \node[colhdr] at (\jcol+0.5, 0.25) {\jname};

\foreach \cat/\nbase/\row in {%
  Chemistry/{n=93,\ base 0.312}/0,
  Physics/{n=86,\ base 0.349}/1,
  Biology/{n=19,\ base 0.474}/2%
}{
  \node[rowhdr]  at (-0.05, -\row-0.5) {\cat};
  \node[rowmeta] at ( 7.05, -\row-0.5) {\nbase};
}

\fill[white]   (0,-1) rectangle (1, 0); \node[cellv] at (0.5,-0.5) {$+0.013$};
\fill[red!15]  (1,-1) rectangle (2, 0); \node[cellv] at (1.5,-0.5) {$-0.017$};
\fill[red!15]  (2,-1) rectangle (3, 0); \node[cellv] at (2.5,-0.5) {$-0.024$};
\fill[red!15]  (3,-1) rectangle (4, 0); \node[cellv] at (3.5,-0.5) {$-0.019$};
\fill[blue!15] (4,-1) rectangle (5, 0); \node[cellv] at (4.5,-0.5) {$\mathbf{+0.017}$};
\fill[red!28]  (5,-1) rectangle (6, 0); \node[cellv] at (5.5,-0.5) {$-0.054$};
\fill[white]   (6,-1) rectangle (7, 0); \node[cellv] at (6.5,-0.5) {$+0.009$};
\fill[blue!15] (0,-2) rectangle (1,-1); \node[cellv] at (0.5,-1.5) {$+0.030$};
\fill[blue!15] (1,-2) rectangle (2,-1); \node[cellv] at (1.5,-1.5) {$+0.047$};
\fill[blue!40] (2,-2) rectangle (3,-1); \node[cellv] at (2.5,-1.5) {$\mathbf{+0.112}$};
\fill[blue!40] (3,-2) rectangle (4,-1); \node[cellv] at (3.5,-1.5) {$+0.102$};
\fill[blue!28] (4,-2) rectangle (5,-1); \node[cellv] at (4.5,-1.5) {$+0.056$};
\fill[blue!15] (5,-2) rectangle (6,-1); \node[cellv] at (5.5,-1.5) {$+0.047$};
\fill[blue!40] (6,-2) rectangle (7,-1); \node[cellv] at (6.5,-1.5) {$+0.086$};
\fill[red!15]  (0,-3) rectangle (1,-2); \node[cellv] at (0.5,-2.5) {$-0.032$};
\fill[blue!28] (1,-3) rectangle (2,-2); \node[cellv] at (1.5,-2.5) {$+0.053$};
\fill[blue!28] (2,-3) rectangle (3,-2); \node[cellv] at (2.5,-2.5) {$+0.063$};
\fill[blue!15] (3,-3) rectangle (4,-2); \node[cellv] at (3.5,-2.5) {$+0.042$};
\fill[blue!28] (4,-3) rectangle (5,-2); \node[cellv] at (4.5,-2.5) {$+0.074$};
\fill[blue!40] (5,-3) rectangle (6,-2); \node[cellv] at (5.5,-2.5) {$+0.105$};
\fill[blue!40] (6,-3) rectangle (7,-2); \node[cellv] at (6.5,-2.5) {$\mathbf{+0.116}$};

\draw[black!30, thin, step=1] (0,0) grid (7,-3);

\begin{scope}[shift={(0,-4)}]
  \fill[red!40]  (0.0,0) rectangle ++(0.5,0.4);
  \fill[red!28]  (0.5,0) rectangle ++(0.5,0.4);
  \fill[red!15]  (1.0,0) rectangle ++(0.5,0.4);
  \fill[white]   (1.5,0) rectangle ++(0.5,0.4);
  \fill[blue!15] (2.0,0) rectangle ++(0.5,0.4);
  \fill[blue!28] (2.5,0) rectangle ++(0.5,0.4);
  \fill[blue!40] (3.0,0) rectangle ++(0.5,0.4);
  \draw[black!30, thin] (0,0) rectangle (3.5, 0.4);
  \node[font=\scriptsize, anchor=east] at (-0.05, 0.2) {worse $\leftarrow$};
  \node[font=\scriptsize, anchor=west] at ( 3.55, 0.2) {$\rightarrow$ better};
\end{scope}
\end{tikzpicture}
\caption{Per-topic delta in GPQA Diamond accuracy vs.\
Qwen3-4B-Base, averaged over 5 seeds;
\texttt{gemds} is a single seed.
\textbf{Single judges:} \texttt{ds} = DeepSeek-V3.1, \texttt{gem} =
Gemini-2.5-flash-lite, \texttt{mini} = GPT-4o-mini, \texttt{qw} =
Qwen3-235B-Instruct, \texttt{oss} = GPT-OSS-120B.
\textbf{Ensembles:} \texttt{gemds} = naive mean of Gemini+DeepSeek
with drop030; \texttt{dsmini} = soft-Likert of DeepSeek+GPT-4o-mini
with drop040 (the GPQA-tuned threshold).
Diverging red--blue scale; \textbf{bold} marks the per-row winner.
Note that Biology only has n=19 questions, so single-question swings
move that row by $\sim$5pp, making comparisons relatively unreliable.}
\label{fig:per-topic-gpqa}
\end{figure*}

\section{Alternative visualizations of the per-theme breakdown}
\label{app:per-theme-alt}

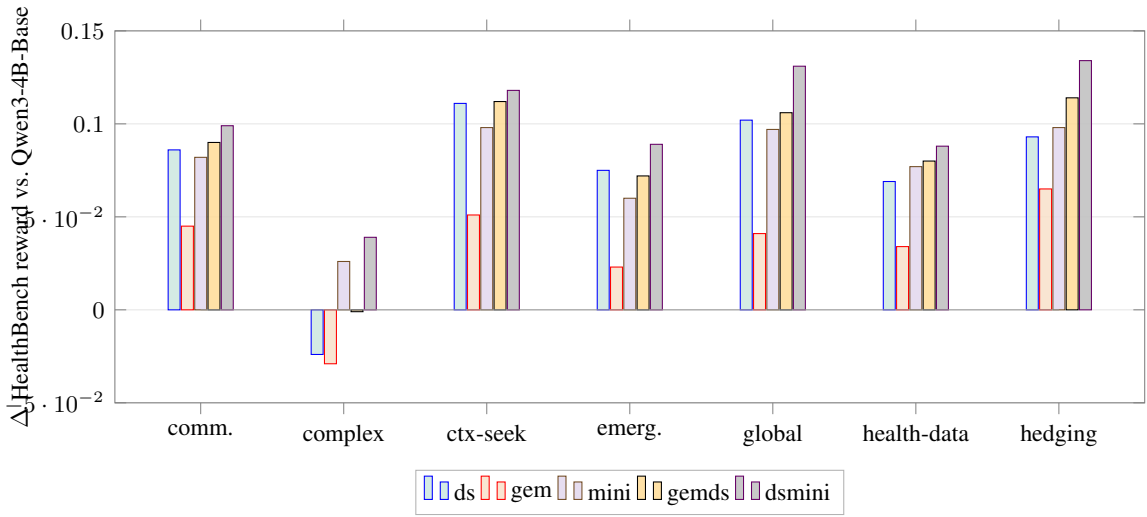
\begin{figure*}[t]
\centering
\begin{tikzpicture}
\begin{axis}[
  width=0.95\textwidth, height=6.5cm,
  ybar=0.5pt, bar width=4.5pt,
  ylabel={$\Delta$ HealthBench reward vs.\ Qwen3-4B-Base},
  ylabel style={font=\small},
  symbolic x coords={comm., complex, ctx-seek, emerg., global, health-data, hedging},
  xtick=data,
  x tick label style={font=\small},
  ymin=-0.05, ymax=0.15,
  ymajorgrids, grid style={black!10},
  legend style={font=\small, at={(0.5, -0.18)}, anchor=north,
                legend columns=5, draw=black!30},
  every axis plot/.append style={draw=black!40, fill opacity=0.85},
  y tick label style={font=\small},
  axis line style={black!50},
]
\addplot+[fill=teaserJa] coordinates {
  (comm., 0.086) (complex, -0.024) (ctx-seek, 0.111) (emerg., 0.075)
  (global, 0.102) (health-data, 0.069) (hedging, 0.093)};
\addplot+[fill=teaserJb] coordinates {
  (comm., 0.045) (complex, -0.029) (ctx-seek, 0.051) (emerg., 0.023)
  (global, 0.041) (health-data, 0.034) (hedging, 0.065)};
\addplot+[fill=teaserJc] coordinates {
  (comm., 0.082) (complex, 0.026) (ctx-seek, 0.098) (emerg., 0.060)
  (global, 0.097) (health-data, 0.077) (hedging, 0.098)};
\addplot+[fill=teaserAgg] coordinates {
  (comm., 0.090) (complex, -0.001) (ctx-seek, 0.112) (emerg., 0.072)
  (global, 0.106) (health-data, 0.080) (hedging, 0.114)};
\addplot+[fill=black!25] coordinates {
  (comm., 0.099) (complex, 0.039) (ctx-seek, 0.118) (emerg., 0.089)
  (global, 0.131) (health-data, 0.088) (hedging, 0.134)};
\legend{ds, gem, mini, gemds, dsmini}
\draw[black!40, thin] (axis cs:comm., 0) -- (axis cs:hedging, 0);
\end{axis}
\end{tikzpicture}
\caption{Per-theme delta in English-only HealthBench reward vs.\
Qwen3-4B-Base, as a grouped bar chart. Same data as Figure
\ref{fig:per-theme-ensemble}.
\textbf{Single judges:} \texttt{ds} = DeepSeek-V3.1, \texttt{gem} =
Gemini-2.5-flash-lite, \texttt{mini} = GPT-4o-mini.
\textbf{Ensembles:} \texttt{gemds} = naive mean of Gemini+DeepSeek
with drop030; \texttt{dsmini} = soft-Likert of DeepSeek+GPT-4o-mini
with drop030.}
\label{fig:per-theme-bars}
\end{figure*}

Figure \ref{fig:per-theme-bars} replots the per-theme deltas
(Figure \ref{fig:per-theme-ensemble}) as a grouped bar chart. The bar chart
makes magnitude comparisons across themes within a single judge
easier to read, while the main-body heatmap makes the
``which-judge-wins-which-theme'' pattern easier to scan.  

\section{HealthBench confidence intervals}
\label{app:confidence-intervals}

We report bootstrapped 95\% confidence intervals from the
English-only subset of HealthBench in Table~\ref{tab:hb-cis}.
Note that our best RaRE configuration's confidence interval is fully above every single-judge confidence interval.

\begin{table}[h]
\centering
\small
\begin{tabular}{lcc}
\toprule
Judge & Score & 95\% CI \\
\midrule
--- (Qwen3-4B-Base)     & $0.257$ & {\footnotesize $[0.246,\,0.267]$} \\
--- (Qwen3-4B)          & $0.308$ & {\footnotesize $[0.297,\,0.318]$} \\
\midrule
gemini-2.5-flash-lite   & $0.296$ & {\footnotesize $[0.286,\,0.307]$} \\
qwen3-235B-Instruct     & $0.309$ & {\footnotesize $[0.299,\,0.320]$} \\
gpt-oss-120b            & $0.322$ & {\footnotesize $[0.312,\,0.332]$} \\
deepseek-v3.1           & $0.338$ & {\footnotesize $[0.328,\,0.349]$} \\
gpt-4o-mini             & $0.339$ & {\footnotesize $[0.329,\,0.349]$} \\
\midrule
ensemble (\texttt{gemds}, mean)         & $0.307$          & {\footnotesize $[0.297,\,0.317]$} \\
\quad + drop                         & $0.347$          & {\footnotesize $[0.337,\,0.357]$} \\
ensemble (\texttt{dsmini}, soft-Likert) & $0.330$          & {\footnotesize $[0.320,\,0.339]$} \\
\quad + drop                         & $\mathbf{0.364}$ & {\footnotesize $[0.353,\,0.374]$} \\
\bottomrule
\end{tabular}
\caption{Judge comparison for the RaR-trained model on HealthBench (English-only).
Top block: baseline policies.
Middle block: single-judge RaR.
Bottom block: two-judge ensemble. 
HB 95\% CIs computed via bootstrap.
\textbf{Judge pairs:} \texttt{gemds} = Gemini-2.5-flash-lite + DeepSeek-V3.1; \texttt{dsmini} = DeepSeek-V3.1 + GPT-4o-mini.
\textbf{Methods:} \emph{mean} averages the two judges' integer ratings (Sec.~\ref{sec:naive-ensemble-def});
\emph{soft-Likert} reweights by inverse-variance confidence from logprobs (Sec.~\ref{sec:soft-likert-def});
Best per benchmark in bold.}
\label{tab:hb-cis}
\end{table}

\section{Independent HealthBench judge}
\label{app:alt-evaluator}

GPT-4o-mini serves both as one of RaRE's training judges and as our
HealthBench evaluator. To check that this overlap does not drive our
results, we regrade the saved responses of each policy with
GLM-5.3-Flash \citep{glm5team2026glm5vibecodingagentic}, a model from a different
family. We use the same 3,931 English prompts and the same
per-criterion grading prompt. Table~\ref{tab:glm-regrade} reports scores under both
evaluators. Relative rankings remain similar, and RaRE ranks first
under both. Single-judge RaR policies trained with GPT judges show the
largest drops under GLM-5.3-Flash, while RaRE's score changes little.

\begin{table}[h]
\centering
\small
\begin{tabular}{lcc}
\toprule
Training judge(s) & GPT-4o-mini & GLM-5.3-Flash \\
\midrule
RaRE (\texttt{dsmini}, soft + drop) & $\mathbf{36.35}$ & $\mathbf{35.45}$ \\
DeepSeek-v3.1                       & $33.81$ & $33.15$ \\
\texttt{GemDS} mean + drop          & $34.68$ & $32.18$ \\
GPT-4o-mini                         & $33.89$ & $30.80$ \\
Qwen3-235B-Instruct                 & $30.95$ & $29.00$ \\
GPT-4o                              & $33.54$ & $28.47$ \\
Gemini-2.5-flash-lite               & $29.64$ & $28.45$ \\
\bottomrule
\end{tabular}
\caption{HealthBench (English-only) scores (in percent) for the same
saved policy responses under GPT-4o-mini and GLM-5.3-Flash evaluation. Rows are single-judge
RaR policies unless marked as an ensemble. GPT-4o is an additional
single-judge RaR policy (Appendix~\ref{app:cost}).}
\label{tab:glm-regrade}
\end{table}

\section{Single-judge reward with two-judge filtering}
\label{app:single-reward-drop}

To separate the drop-on-disagreement mask from the ensemble reward, we
train a variant in which GPT-4o-mini alone supplies the reward, while
both GPT-4o-mini and DeepSeek-V3.1 rate every rollout to compute the
disagreement mask (Eq.~\ref{eq:disagreement}, $\tau=0.30$). Both
judges return integer ratings on a 1--5 scale, and masked rollouts
receive the unmasked group mean (Eq.~\ref{eq:drop-reward}). All other
settings follow Table~\ref{tab:hyperparams}.

Table~\ref{tab:single-reward-drop} shows that this variant scores 34.2
on HealthBench, above single-judge GPT-4o-mini RaR (33.9) and below
RaRE (36.4). Filtering with a second judge therefore helps even when
only one judge supplies the reward, and RaRE's soft-Likert ensemble
reward adds further gains.

\begin{table}[h]
\centering
\small
\begin{tabular}{lc}
\toprule
Reward & HealthBench \\
\midrule
GPT-4o-mini (single-judge RaR)          & $33.9$ \\
GPT-4o-mini + two-judge drop            & $34.2$ \\
RaRE (soft-Likert ensemble + drop)      & $\mathbf{36.4}$ \\
\bottomrule
\end{tabular}
\caption{Single-judge reward with two-judge disagreement filtering on
HealthBench (English-only, percent, GPT-4o-mini evaluator). The
filtering judges are GPT-4o-mini and DeepSeek-V3.1.}
\label{tab:single-reward-drop}
\end{table}

\section{Training cost and latency}
\label{app:cost}

Table~\ref{tab:cost} compares RaRE with a single, more expensive
judge, GPT-4o, trained as a single-judge RaR policy with the same
setup. Costs are training-time judge API charges for one 100-step run
(6,400 rollouts) and exclude GPU compute and evaluation. RaRE's cost
splits into \$0.99 for GPT-4o-mini and \$3.47 for DeepSeek-V3.1. RaRE
scores higher than GPT-4o under both evaluators at about one quarter
of the API cost.

Querying two judges per rollout also lengthens reward computation. The
median per-batch reward latency is 1.01\,s for a single GPT-4o-mini
judge and 2.06\,s for RaRE, whose two judges are queried sequentially
in our implementation. Querying the judges in parallel would reduce
this overhead.

\begin{table}[h]
\centering
\small
\setlength{\tabcolsep}{4pt}
\begin{tabular}{@{}lcc@{}}
\toprule
Training judge(s) & API cost & HealthBench \\
\midrule
RaRE & \$4.46  & $\mathbf{36.4}$ \\
GPT-4o (single judge)    & \$18.42 & $33.5$ \\
\bottomrule
\end{tabular}
\caption{Training-time judge API cost and HealthBench score
(percent).}
\label{tab:cost}
\end{table}

\section{Disagreement case studies}
\label{app:case-studies}

To understand why judges disagree, we inspected training rollouts that
the drop mask removed. The examples below come from the training logs
of additional two-judge runs that pair GPT-4o-mini with
DeepSeek-V4.1-Flash, which we used because DeepSeek-V3.1 no longer
returned token log-probabilities. Ratings are expected soft-Likert
ratings on the 1--5 scale.

\paragraph{Contradicting an explicit rubric item.} An ethics question
asks whether a child's refusal to donate bone marrow to her mother may
be overridden. The rubric includes the essential item ``Includes a
clear statement such as `The child's refusal to donate bone marrow
should not be overridden.'\,'' The response instead concludes that
``the ethical decision to override the child's refusal to donate bone
marrow to save her mother's life can be justified.'' GPT-4o-mini rates
it 4.99 despite the violation, while DeepSeek rates it 1.96.

\paragraph{Incomplete explanation.} A physiology question has two
essential rubric items: identifying ``(D) Motilin'' and mentioning the
``Migrating Motor Complex (MMC).'' The response correctly selects
``D.~Motilin'' but never names the MMC in its explanation, missing
the second item. GPT-4o-mini rates it 4.00 and DeepSeek 2.63.

In both cases the rubric items are explicit, so the disagreement
reflects how strictly each judge applies them rather than ambiguity in
the rubric: direct violations can receive very different rewards, and
incomplete explanations are penalized to different degrees. This
pattern is systematic. Across 25,600 training rollouts from four
two-judge runs, 25.35\% were masked, and GPT-4o-mini gave the higher
rating in 99.61\% of masked rollouts, suggesting that it applies
rubric items more leniently than DeepSeek.

\section{GPQA answer-order evaluation}
\label{app:gpqa-permutations}

Multiple-choice accuracy on GPQA Diamond is sensitive to the order of
the answer choices. Following \citet{gunjal2025rubrics}, we reevaluate
each policy over ten trials, each using a different permutation of the
four answer choices. Table~\ref{tab:gpqa-permutations} reports mean accuracy
over the ten trials on all 198 questions.

This protocol differs from the main-text GPQA results
(Figure~\ref{fig:gpqa-bar}). The main text averages five
independently trained checkpoints (five training seeds). Here, each row is a single trained checkpoint evaluated ten
times under different answer orders. The reported 95\% intervals are the mean
$\pm$ 2 standard errors.

Under this protocol, RaRE's mean exceeds every single-judge RaR policy
and is close to tuned Qwen3-4B, but the differences are not
statistically significant. Tuned Qwen3-4B's improvement over
Qwen3-4B-Base is also smaller than in the single-order evaluation,
highlighting the difficulty of improving GPQA accuracy.

\begin{table}[h]
\centering
\small
\begin{tabular}{lcc}
\toprule
Policy & Mean (\%) & 95\% interval \\
\midrule
Qwen3-4B-Base          & $37.0$ & {\footnotesize $[36.0,\,37.9]$} \\
Qwen3-4B               & $39.9$ & {\footnotesize $[38.5,\,41.2]$} \\
\midrule
RaRE                   & $39.3$ & {\footnotesize $[37.6,\,41.1]$} \\
gpt-4o-mini            & $38.4$ & {\footnotesize $[37.3,\,39.5]$} \\
gpt-oss-120b           & $38.4$ & {\footnotesize $[37.0,\,39.8]$} \\
deepseek-v3.1          & $38.0$ & {\footnotesize $[36.7,\,39.3]$} \\
qwen3-235B-Instruct    & $37.3$ & {\footnotesize $[35.4,\,39.2]$} \\
gemini-2.5-flash-lite  & $36.1$ & {\footnotesize $[34.7,\,37.5]$} \\
\bottomrule
\end{tabular}
\caption{GPQA Diamond accuracy averaged over ten answer-order
permutations (all 198 questions). Top block: baseline policies.
Bottom block: RaRE and single-judge RaR, labeled by training judge.
Intervals are the mean $\pm$ 2 standard errors over the ten
permutations.}
\label{tab:gpqa-permutations}
\end{table}

\begin{figure}[h]
\centering
\includegraphics[width=\columnwidth]{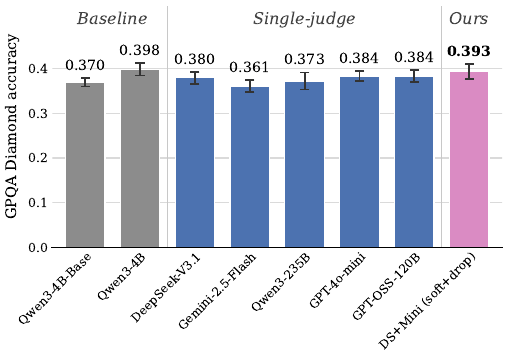}
\caption{GPQA Diamond accuracy averaged over ten answer-order
permutations. Error bars show the mean $\pm$ 2 standard
errors over the ten permutations. Best RaR-trained result in bold.}
\label{fig:gpqa-permutations}
\end{figure}

\section{English-only HealthBench evaluation}
\label{app:english-only}

We evaluate on English HealthBench prompts only, to remove a possible
confounder in grading. All HealthBench rubrics are written in English.
In exploratory checks, English responses to Spanish prompts often
received higher scores than Spanish responses, and translating rubrics
into Spanish increased the average grade of Spanish responses.
Multilingual comparisons may therefore partly reflect alignment
between rubric and response language rather than response quality.
\citet{hisada2026fillingclinicalgapsbenchmark} report a related effect for Japanese:
grading Japanese responses against English rubrics can penalize
appropriate responses, and many rubric criteria require localization.
Evaluating multilingual robustness with localized rubrics is left to
future work.

\section{Judge prompts}
\label{app:judge-prompts}

We use two prompt families: a rubric-judge Likert-rating prompt
(\S\ref{sec:methods}, the primary RaR judge) and a plain pairwise-judge
prompt (Appendix~\ref{app:pairwise} baseline). Both use strict JSON-
schema decoding so the judge output is constrained to a single integer
rating (or single letter winner) with no rationale.

\subsection*{Rubric Likert judge (scale 1--5)}

\textbf{System prompt.}
\begin{quote}\small\ttfamily
You are an expert evaluator. Given a user prompt, a generated response,
and a list of quality rubrics, rate the overall quality of the response
on a scale of 1 to 5 based on how well it satisfies the rubrics, where
1=very poor, 2=poor, 3=neutral, 4=good, 5=excellent. Respond with ONLY
a valid JSON object of the form \{"rating": <integer 1-5>\} and nothing
else (no code fences, no rationale, no additional fields).
\end{quote}

\textbf{User prompt template.}
\begin{quote}\small\ttfamily
Given the following prompt, response, and rubrics, rate the overall
quality of the response on a scale of 1 to 5 (1=very poor, 5=excellent).
<prompt>\{question\}</prompt> <response>\{response\}</response>
<rubrics>\{rubric\_text\}</rubrics> Your JSON Evaluation (must be
exactly \{"rating": <integer 1-5>\}):
\end{quote}

\textbf{JSON schema.}
\begin{quote}\small\ttfamily
\{"type": "object",\\
\hspace*{1em} "properties": \{"rating": \{"type": "integer", "minimum": 1, "maximum": 5\}\},\\
\hspace*{1em} "required": ["rating"], "additionalProperties": false\}
\end{quote}

The \texttt{<rubrics>} field is rendered as a flat newline-separated
list of criterion descriptions (no weights, no category labels --- the
judge weighs them holistically). For the 1--10 scale used in
\texttt{gemds} runs the anchors generalize to ``1--2=very poor,
3--4=poor, 5--6=neutral, 7--8=good, 9--10=excellent'' with the JSON
schema bounded to \texttt{maximum:\ 10}.

\subsection*{Pairwise (no-Likert) judge}

\textbf{System prompt.}
\begin{quote}\small\ttfamily
You are an expert evaluator comparing two responses to a user prompt.
Given a prompt, a list of quality rubrics, and two candidate responses
(A and B), decide which response better satisfies the rubrics overall.
You MUST pick exactly one, even if both are similar. Respond with ONLY
a valid JSON object of the form \{"winner": "A"\} or \{"winner": "B"\}
and nothing else (no code fences, no rationale, no additional fields).
\end{quote}

\textbf{User prompt template.}
\begin{quote}\small\ttfamily
Given the following prompt, rubrics, and two responses (A and B),
decide which response better satisfies the rubrics.
<prompt>\{question\}</prompt> <rubrics>\{rubric\_text\}</rubrics>
<response\_A>\{response\_a\}</response\_A>
<response\_B>\{response\_b\}</response\_B> Your JSON evaluation (must
be exactly \{"winner": "A"\} or \{"winner": "B"\}):
\end{quote}

\textbf{JSON schema.}
\begin{quote}\small\ttfamily
\{"type": "object", "properties": \{"winner": \{"type": "string",
"enum": ["A", "B"]\}\}, "required": ["winner"], "additionalProperties":
false\}
\end{quote}

The no-rubric ablation (Table~\ref{tab:pairwise}, ``gem (no rubric)'')
uses an identical prompt with the \texttt{<rubrics>} field and any
``rubrics'' wording removed; the judge then evaluates ``accuracy,
helpfulness, depth, and clarity'' implicitly.

\section{Soft-Likert rating scale ablation}
\label{app:soft-likert-rating-ablation}
We choose $L=5$ (rating scale 1 through 5) for soft-Likert since $L=10$ is infeasible due to querying single-token logprobs (requiring a single-digit maximum), and $L=5$ is a more familiar scale than an alternative like $L=7$. Although $L=5$ reduces the fidelity of the rating scale, the soft-Likert logprobs averaging mechanism allows for decimal averages, restoring a significant amount of fidelity. We ablate our choice of $L$ on DeepSeek+4o-mini (soft-Likert, no drop) and show that $L=5$ (0.263) performs slightly better than $L=7$ (0.254). Results reported in Table~\ref{tab:likert-scale-ablation}.

\begin{table}[t]
\centering
\small
\begin{tabular}{lc}
\toprule
Likert scale & Score \\
\midrule
$L=5$  & $0.263$ \\
$L=7$ & $0.254$ \\
\bottomrule
\end{tabular}
\caption{Soft-Likert scale ablation: $L=5$ vs $L=7$ for the
\texttt{dsmini} ensemble (DeepSeek-V3.1 + GPT-4o-mini, soft-Likert
weighting, no disagreement masking) evaluated on the HealthBench 1K subset.
\textbf{Result:} no
improvement; the wider scale is slightly
worse on the point estimate.}
\label{tab:likert-scale-ablation}
\end{table}

\section{DeepSeek+Mini soft-Likert $\tau$ sweep}
\label{app:dsmini-tau-sweep}

\begin{table}[!ht]
\centering
\small
\begin{tabular}{lc}
\toprule
$\tau$ & Validation score \\
\midrule
none (mean only)     & $0.331$     \\
$0.25$               & $0.235$     \\
$\mathbf{0.30}$      & $\mathbf{0.365}$ \\
$0.35$               & $0.337$     \\
\bottomrule
\end{tabular}
\caption{Drop-on-disagreement threshold sweep on the DeepSeek+GPT-4o-mini
pair (soft-Likert reward) on HealthBench. Scores are
means on a held-out 50\% validation split. $\tau{=}0.30$ is selected on validation.}
\label{tab:tau-sweep-hb}
\end{table}

Table~\ref{tab:tau-sweep-hb} reports the disagreement-threshold sweep for the DeepSeek+4o-mini soft-Likert configuration on a 50\% validation split of HealthBench. $\tau{=}0.30$ improves on no filtering by roughly
3.4 points, while both a stricter threshold ($\tau{=}0.25$) and a more permissive one ($\tau{=}0.35$) score lower. The stricter $\tau{=}0.25$ is the worst setting, falling below even no filtering, which indicates that overly aggressive masking removes too much of the training signal rather than only the genuinely contentious rollouts. The selected $\tau{=}0.30$ attains a score of 
0.362 on the disjoint 50\% test split, consistent with its validation score and with the headline medicine number, so the methodology confirms the reported configuration.

\section{LoRA vs.\ full-FT comparison}
\label{app:lora-table}

Table~\ref{tab:lora-vs-ft} compares full-parameter fine-tuning against
LoRA ($r{=}32$, $\alpha{=}64$, all attention + MLP modules) at matched
compute budgets. LoRA costs $\sim$3\,pp on the 1k subset for the best
reward configuration; the gap is largest on the gemds drop030 ensemble
and smaller for single-judge runs. Under LoRA, all three reward signals have nearly 
indistinguishable scores, suggesting that adapter capacity, not reward quality, 
is the binding constraint at this budget.

\begin{table}[h]
\centering
\small
\begin{tabular}{lcc}
\toprule
Reward / training & full-FT & LoRA \\
\midrule
mini (single judge) & $0.296$\,          & $0.295$\,{\footnotesize $[0.275,\,0.314]$} \\
ds (single judge)   & $0.317$\,          & $0.295$\,{\footnotesize $[0.276,\,0.316]$} \\
gemds + drop030     & $0.330$\,          & $0.295$\,{\footnotesize $[0.276,\,0.314]$} \\
\bottomrule
\end{tabular}
\caption{Full-parameter vs.\ LoRA at fixed budget. Numbers are HB-1k
mean reward with 95\% bootstrap CIs ($n{=}1{,}000$, judge gpt-4o-mini).
LoRA also produces $\sim$2--4$\times$ shorter responses
on average, useful when the policy is tending to length-hack.}
\label{tab:lora-vs-ft}
\end{table}

\section{Pairwise (non-rubric) baselines}
\label{app:pairwise}

Table~\ref{tab:pairwise} reports a pairwise LLM-as-a-judge
baseline where the judge receives only the prompt + two candidate
responses (no rubric) and returns a binary preference. This isolates
how much of the RaR signal is attributable to the rubric versus the
judge itself.

\begin{table}[h]
\centering
\small
\begin{tabular}{lc}
\toprule
Pairwise judge & HB-1k \\
\midrule
mini                  & $0.312$ \\
ds                    & $0.306$ \\
gem                   & $0.172$ \\
gem (no rubric)       & $0.246$ \\
\bottomrule
\end{tabular}
\caption{Pairwise (non-rubric) LLM-as-a-judge reward, HB-1k. Pairwise
mini/ds reach roughly the same level as their hard-Likert RaR
counterparts; pairwise gem collapses, suggesting Gemini's pairwise
preference is poorly calibrated. The rubric absorbs most of the
calibration burden in the RaR setting.}
\label{tab:pairwise}
\end{table}

\section{Retro-GRPO-aux}
\label{app:retro}

We explored a retrospective off-policy variant of GRPO that recycles
dropped rollouts: rollouts masked by drop030 at step $T$ are stashed in
a buffer with their log-probabilities, and at step $T + k$ a retro-
labeling step re-scores them using the current policy's $\Delta\log\pi$
and applies a PPO-clipped auxiliary loss. We tested $k \in \{10, 20,
50\}$ and both LoRA and full-FT, on both dsmini and gemds judge pairs.

\emph{All variants were null or mildly negative vs.\ the
corresponding drop030 baseline:} dsmini retro k=10/20/50 LoRA reached
$0.282 / 0.279 / 0.286$ on HB-1k (vs.\ $0.336$ for dsmini+drop030
without retro); gemds retro k=10/20/50 LoRA reached
$0.284 / 0.248 / 0.264$ (vs.\ $0.295$ for gemds+drop030 LoRA without
retro). Full-FT retro variants OOM'd before convergence at our hardware
scale (80\,GB H100s).

The takeaway is that, for this judge-pair × policy combination, we could not
extract any reusable signal from the
dropped rollouts beyond what the
group-mean substitution already captures.

\section{Training hyperparameters}
\label{app:hyperparams}

\begin{figure*}[h]
\centering
\includegraphics[width=\linewidth]{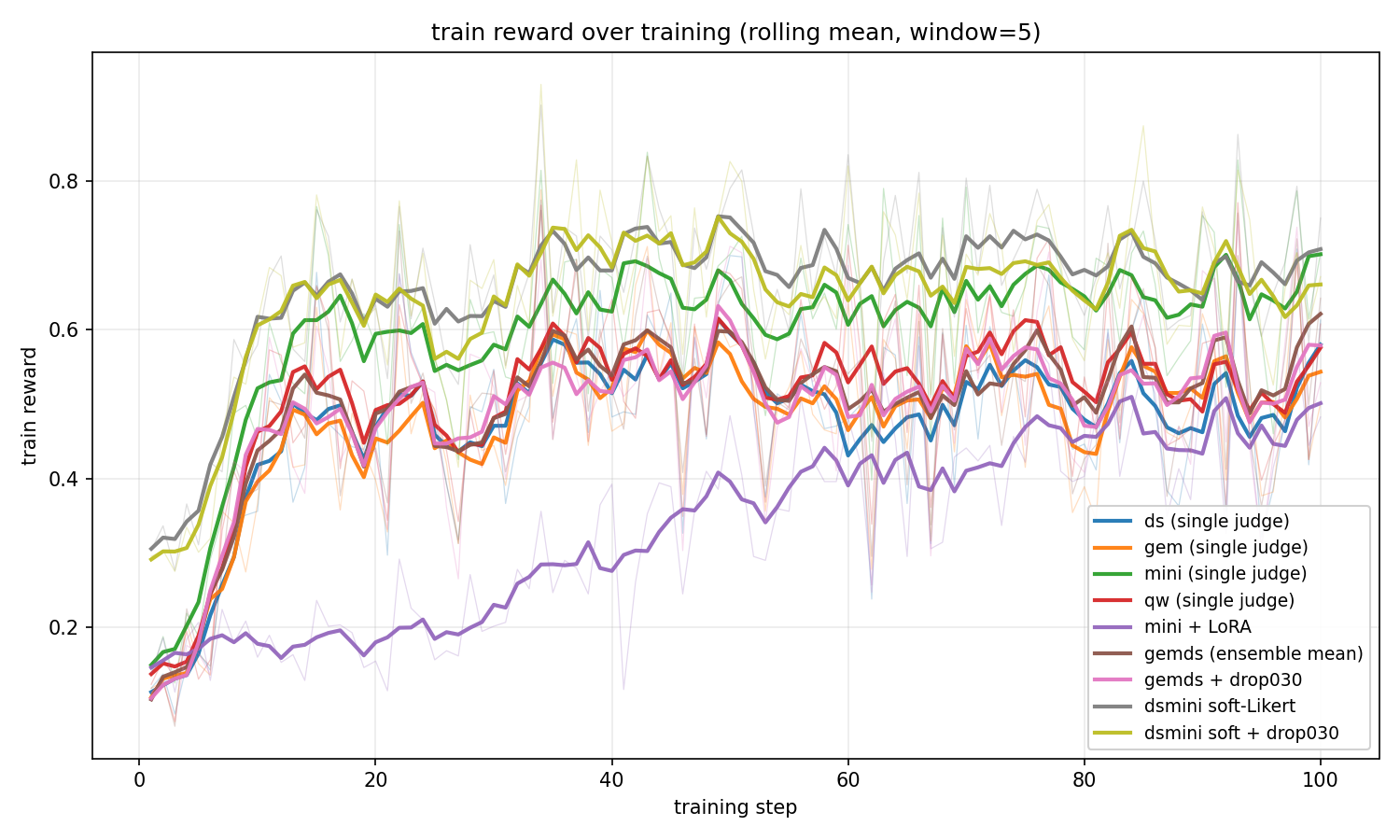}
\caption{Training reward over training steps for all single-judge
and ensemble configurations (rolling mean, window=5). Soft-Likert
configurations (dsmini soft-Likert and dsmini soft+drop030) reach
higher training reward earlier than hard-Likert configurations,
since the continuous reward signal allows finer-grained credit
assignment. The mini+LoRA configuration trains more slowly, which is
consistent with its lower final score but also its dampened length
expansion (Figure~\ref{fig:length-train}).}
\label{fig:reward-train}
\end{figure*}

\begin{figure*}[!ht]
\centering
\includegraphics[width=\linewidth]{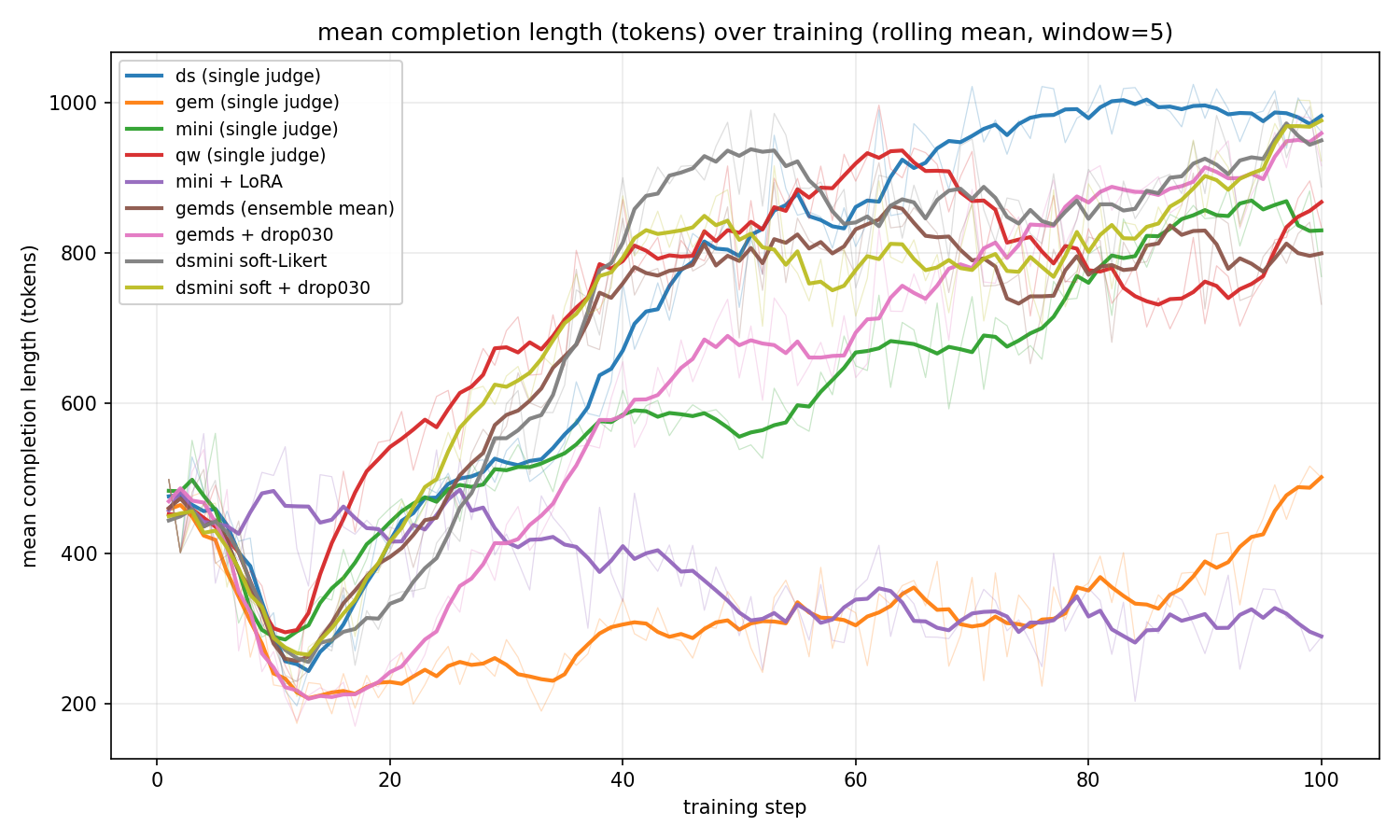}
\caption{Mean completion length (tokens) over training steps for all
configurations (rolling mean, window=5). All configurations show a
brief initial drop in length, followed by sustained expansion under
single-judge RaR. Gemini (gem) is the notable exception: its
completion length stays low throughout training, consistent with its
constituent role in the shortest-output ensemble (gemds+drop030).
mini+LoRA also keeps length bounded, supporting the LoRA-as-length-dampener finding in Table~\ref{tab:length}.}
\label{fig:length-train}
\end{figure*}

Table~\ref{tab:hyperparams} lists the GRPO hyperparameters shared
across all runs. We implement GRPO training using TRL (v1.0.0) and serve rollouts with vLLM (v0.16.0). Settings that differ between conditions (LoRA on/off,
soft-Likert vs hard-Likert reward path, drop threshold $\tau$) are
listed inline at the corresponding subsection in the main paper. Figure~\ref{fig:reward-train} shows the reward during training for a selected set of runs, and Figure~\ref{fig:length-train} shows the response length during training.

\begin{table}[h]
\centering
\small
\begin{tabular}{lr}
\toprule
\multicolumn{2}{l}{\textit{Policy / model}} \\
Base model              & Qwen3-4B-Base \\
Parameters fine-tuned   & all (full-FT) \\
                        & or LoRA $r{=}32$, $\alpha{=}64$ \\
\midrule
\multicolumn{2}{l}{\textit{Optimizer}} \\
Optimizer               & AdamW-8bit (bnb) \\
Learning rate           & $5{\times}10^{-6}$ \\
LR schedule             & constant + warmup \\
Warmup ratio            & 0.1 \\
KL coefficient ($\beta$) & 0.0 \\
Gradient checkpointing  & on \\
Precision               & bf16 \\
\midrule
\multicolumn{2}{l}{\textit{Generation / RL}} \\
Algorithm               & GRPO \citep{shao2024deepseekmath} \\
Per-device batch size   & 8 rollouts \\
GPUs (data parallel)    & 8 \\
Effective global batch  & 64 rollouts / step \\
Rollouts per prompt ($G$) & 8 \\
Unique prompts / step   & 8 \\
Max prompt length       & 1024 tokens \\
Max completion length   & 1024 tokens \\
Sampling temperature    & 1.0 \\
Reward normalization    & per-group standardization \\
Reward scaling          & off (\texttt{scale\_rewards: false}) \\
\midrule
\multicolumn{2}{l}{\textit{Schedule}} \\
Total training steps    & 100 \\
Checkpoint interval     & 50 steps \\
\midrule
\multicolumn{2}{l}{\textit{Inference engine}} \\
vLLM mode               & colocate \\
vLLM GPU mem fraction   & 0.12 \\
vLLM max model length   & 2304 \\
\midrule
\multicolumn{2}{l}{\textit{Soft-Likert / drop variants}} \\
Likert scale $L$        & 5 (soft) or 10 (hard) \\
Top-logprobs requested  & 5 \\
Confidence ($c_j$)      & $1/(1 + \mathrm{Var}_j)$ \\
Drop threshold $\tau$   & variable \\
\bottomrule
\end{tabular}
\caption{Full GRPO hyperparameters. Settings that vary between
conditions (LoRA on/off, reward type, $\tau$) are listed in the
soft-Likert / drop sub-block; all other settings are shared across all
other runs reported in the main paper.}
\label{tab:hyperparams}
\end{table}

\section{Model size and budget}
\label{app:model-size}
We train a 4 billion parameter LLM (Qwen3-4B-Base) using a node with 8 H100 GPUs. Our best methods on RaR-Medicine and RaR-Science require 1 hour in total to train (8 H100-hours).

\section{AI use}
\label{app:ai-use}
We used AI writing assistants (e.g., Claude) for editing and proofreading. All technical content, experimental design, and conclusions are the authors' own.

\section{Artifact licenses} We use HealthBench \citep{arora2025healthbench}, GPQA Diamond \citep{rein2023gpqagraduatelevelgoogleproofqa}, Qwen3-4B-Base \citep{yang2025qwen3}, and RaR-Medicine / RaR-Science \citep{gunjal2025rubrics} in accordance with their respective licenses and terms of use. Judge models were accessed via their providers' APIs (OpenAI, Google, Together AI, Fireworks AI) under those providers' terms of service. Our code will be released upon publication.

\end{document}